%% file: main.tex
\documentclass[10pt,journal,compsoc]{IEEEtran}

\usepackage[T1]{fontenc}
\usepackage[pagebackref=true,breaklinks=true,colorlinks,citecolor=blue,urlcolor=blue,linkcolor=blue]{hyperref}
\usepackage[numbers]{natbib}
\usepackage{amsmath,amssymb,amsfonts}
\usepackage{bbm}
\usepackage{algorithm}
\usepackage{algorithmic}
\usepackage{array}
\usepackage{multirow}
\usepackage{booktabs}
\usepackage{longtable}
\usepackage{tablefootnote}
\usepackage{colortbl}
\usepackage{graphicx}
\usepackage{adjustbox}
\usepackage{sidecap}
\usepackage{textcomp}
\usepackage{stfloats}
\usepackage{url}
\usepackage{verbatim}
\usepackage{xcolor}
\usepackage{xspace}
\usepackage{chngcntr}
\usepackage{inconsolata}
\usepackage{listings}
\usepackage{xr}
\usepackage{tcolorbox}
\usepackage{threeparttable}
\usepackage{lmodern}
\usepackage{paralist}
\usepackage{multicol}
\usepackage{pifont}
\usepackage{subcaption}
\usepackage{fontawesome5} 
\usepackage{xcolor}      

\ifCLASSINFOpdf
\else
\fi

\definecolor{mygray}{gray}{0.9}
\begin{document}

\input{preamble}
\title{LLM Post-Training: A Deep Dive into Reasoning\\ Large Language Models}

\author{Komal Kumar$^*$, Tajamul Ashraf$^*$, Omkar Thawakar, Rao Muhammad Anwer, Hisham Cholakkal, \\ Mubarak Shah, 
Ming-Hsuan Yang, Phillip H.S. Torr, Fahad Shahbaz Khan, Salman Khan \\
\IEEEcompsocitemizethanks{
\IEEEcompsocthanksitem 
$^*$Equal contribution. 
Corresponding authors (Email: komal.kumar@mbzuai.ac.ae, tajamul.ashraf@mbzuai.ac.ae)
\IEEEcompsocthanksitem Komal Kumar, Tajamul Ashraf, Omkar Thawakar, Rao Muhammad Anwer, Hisham Cholakkal, Fahad Shahbaz Khan, and Salman Khan are with Mohamed bin Zayed University of Artificial Intelligence, Abu Dhabi, UAE.
\IEEEcompsocthanksitem Mubarak Shah is with the Center for Research in Computer Vision at the University of Central Florida, Orlando, FL 32816, USA.
\IEEEcompsocthanksitem Ming-Hsuan Yang is with the University of California at Merced, Merced,
CA 95343 USA, and also with Google DeepMind, Mountain View, CA 94043, USA.
\IEEEcompsocthanksitem Philip H.S. Torr is with the Department of Engineering
Science, University of Oxford, Oxford OX1 2JD, UK.
}}

\markboth{}%
{Shell \MakeLowercase{\textit{et al.}}: Bare Demo of IEEEtran.cls for IEEE Journals}

\input{content/abstract}
\maketitle
\IEEEdisplaynontitleabstractindextext
\IEEEpeerreviewmaketitle

\input{content/introduction}
\input{content/background}
\input{content/methods}

\input{content/datasets}
\input{content/directions}
\input{content/conclusion}
\appendices

\ifCLASSOPTIONcaptionsoff
  \newpage
\fi

\bibliographystyle{ieeetr}  
{\footnotesize
\bibliography{main}}  %

\end{document}

%% file: preamble.tex
\def\instructgpt{\texttt{Instruct-GPT}\xspace}
\def\gptfour{\texttt{GPT-4}\xspace}
\def\gemini{\texttt{Gemini}\xspace}
\def\internlm{\texttt{InternLM2}\xspace}
\def\claude{\texttt{Claude 3}\xspace}
\def\reka{\texttt{Reka}\xspace}
\def\zephyr{\texttt{Zephyr}\xspace}
\def\deepseek{\texttt{DeepSeek-V2}\xspace}
\def\chatglm{\texttt{ChatGLM}\xspace}
\def\nemotron{\texttt{Nemotron-4 340B}\xspace}
\def\llama{\texttt{Llama 3}\xspace}
\def\qwen{\texttt{Qwen2}\xspace}
\def\gemma{\texttt{Gemma2}\xspace}
\def\starling{\texttt{Starling-7B}\xspace}
\def\athene{\texttt{Athene-70B}\xspace}
\def\hermes{\texttt{Hermes 3}\xspace}
\def\oone{\texttt{o1}\xspace}
\def\llms{\texttt{LLMs}\xspace}
\def\llm{\texttt{LLM}\xspace}
\def\rl{\texttt{RL}\xspace}
\def\rlhf{\texttt{RLHF}\xspace}
\def\rag{\texttt{RAG}\xspace}
\def\mdp{\texttt{MDP}\xspace}
\def\grpo{\texttt{GRPO}\xspace}
\def\trpo{\texttt{TRPO}\xspace}
\def\dpo{\texttt{DPO}\xspace}
\def\mle{\texttt{MLE}\xspace}
\def\rag{\texttt{RAG}\xspace}
\def\reinforce{\texttt{REINFORCE}\xspace}
\def\ppo{\texttt{PPO}\xspace}
\def\trpo{\texttt{TRPO}\xspace}
\def\mrt{\texttt{MRT}\xspace}
\def\seqgan{\texttt{SeqGAN}\xspace}
\def\mixer{\texttt{MIXER}\xspace}
\def\lora{\texttt{LoRA}\xspace}
\def\cot{\texttt{CoT}\xspace}
\def\rlaif{\texttt{RLAIF}\xspace}
\def\scst{\texttt{SCST}\xspace}

\newcommand{\insightbox}[1]{%
    \begin{tcolorbox}[colframe=black!70, colback=blue!5, boxrule=1pt, arc=4mm]
        \includegraphics[width=0.4cm]{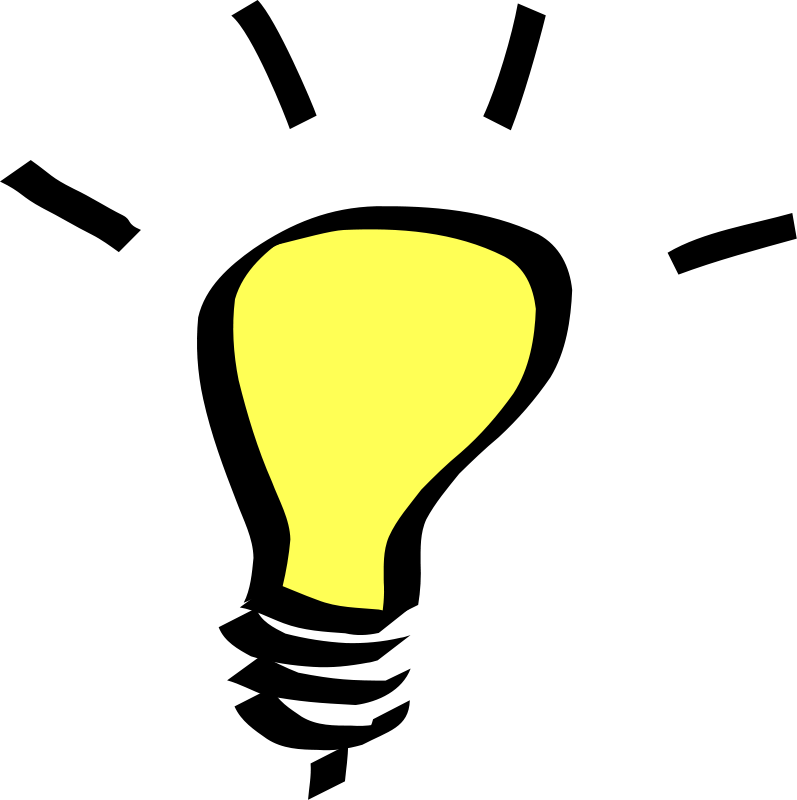}
        \textbf{\small#1}
    \end{tcolorbox}
}

\newcommand{\warningbox}[1]{%
    \begin{tcolorbox}[colframe=red!70, colback=yellow!10, boxrule=1pt, arc=4mm]
        \faExclamationTriangle\ \textbf{#1}
    \end{tcolorbox}
}

%% file: content/abstract.tex
\IEEEtitleabstractindextext{
\begin{abstract}
Large Language Models (\llms) have transformed the natural language processing landscape and brought to life diverse applications. 
Pretraining on vast web-scale data has laid the foundation for these models, yet the research community is now increasingly shifting focus toward post-training techniques to achieve further breakthroughs.
While pretraining provides a broad linguistic foundation, post-training methods enable \llms to refine their knowledge, improve reasoning, enhance factual accuracy, and align more effectively with user intents and ethical considerations.
Fine-tuning, reinforcement learning, and test-time scaling have emerged as critical strategies for optimizing \llms performance, ensuring robustness, and improving adaptability across various real-world tasks. 
This survey provides a systematic exploration of post-training methodologies, analyzing their role in refining LLMs beyond pretraining, addressing key challenges such as catastrophic forgetting, reward hacking, and inference-time trade-offs. 
We highlight emerging directions in model alignment, scalable adaptation, and inference-time reasoning, and outline future research directions. 
We also provide a public repository to continually track developments in this fast-evolving field:  
\textcolor{purple}{\url{https://github.com/mbzuai-oryx/Awesome-LLM-Post-training}}.

\end{abstract}
}

%% file: content/introduction.tex
\section{Introduction}
\IEEEPARstart{C}{ontemporary} Large Language Models (\llms) exhibit remarkable capabilities across a vast spectrum of tasks, encompassing not only text generation \cite{vaswani2017attention, brown2020language, yang2019xlnet} and question-answering~\cite{devlin2018bert, lan2019albert, raffel2020exploring, verga2024replacing},  but also sophisticated multi-step reasoning~\cite{wei2022chain, wang2024q, wu2024mindmap,ding2024crosscodeeval}.
They power applications in natural language understanding~\cite{achiam2023gpt,dubey2024llama,team2024gemma,team2023gemini,liu2024deepseek,abdin2024phi}, content generation~\cite{fan2018hierarchical,chhun2022human,arif2024fellowship,ye2023selfee,musolesi2024creative,ren2023self,lewis2020retrieval,lai2023ds}, automated reasoning~\cite{zhu2024starling,paul2023refiner,xie2024self,ye2024self}, and multimodal interactions~\cite{luo2024videoautoarena,deng2024efficient, ashraf2024fate,chen2024mllm}. By leveraging vast self-supervised training corpora, these models often approximate human-like cognition~\cite{hagendorff2023human,pan2024human,chen2024humans,newell1972human,wang2024human}, demonstrating impressive adaptability in real-world settings.

\begin{figure}[t]
    \centering
    \includegraphics[width=\linewidth]{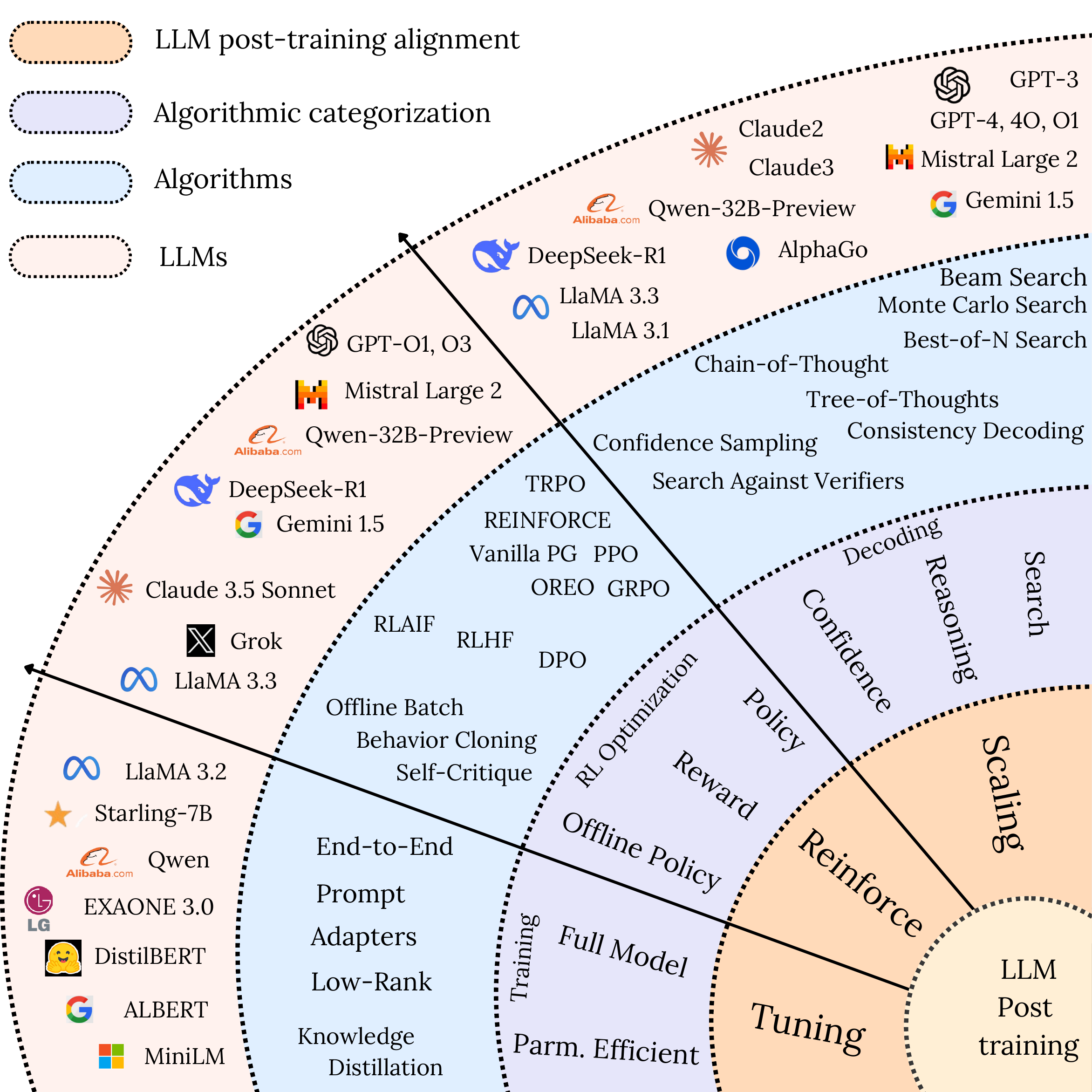}
    \put(-65,40){\scriptsize{\S\ \ref{sec_reinforcce}}} 
    \put(-73,5){\scriptsize{\S\ \ref{sec_fintune}}}  
    \put(-35,65){\scriptsize{\S\ \ref{sec_tts}}}  
     \caption{A taxonomy of post-training approaches for \llms (\llms), categorized into Fine-tuning, Reinforcement Learning, and Test-time Scaling methods. We summarize the key techniques used in recent \llm models, such as GPT-4~\cite{openai2023gpt}, LLaMA 3.3~\cite{dubey2024llama}, and Deepseek R1~\cite{guo2025deepseek}.}
    \label{fig:intro_main}
\end{figure}

Despite these impressive achievements, \llms remain prone to critical shortcomings. They can generate misleading or factually incorrect content (commonly referred to as ``\textit{hallucinations}'') and may struggle to maintain logical consistency throughout extended discourse~\cite{hicks2024chatgpt, maleki2024aihallucinationsmisnomerworth,bruno2023insights,farquhar2024detecting,leiser2024hill,gunjal2024detecting}. {Moreover, the concept of reasoning in \llms remains a topic of debate. While these models can produce responses that appear logically coherent, their reasoning is fundamentally distinct from human-like logical inference~\cite{hao2023reasoning, hagendorff2023human, ye2025limoreasoning, li2025imaginereasoningspacemultimodal}. This distinction is crucial, as it helps explain why \llms can produce compelling outputs while still stumbling on relatively simple logical tasks. Unlike symbolic reasoning that manipulates explicit rules and facts, \llms operate in an implicit and probabilistic manner~\cite{xia2024language, maleki2024aihallucinationsmisnomerworth, ji2023survey}. For the scope of this work, `\emph{reasoning}' in \llms refers to their ability to generate logically coherent responses based on statistical patterns in data rather than explicit logical inference or symbolic manipulation.}
Additionally, models trained purely via next-token prediction can fail to align with user expectations or ethical standards, especially in ambiguous or malicious scenarios~\cite{devlin2018bert,he2024lawnexttokenpredictionlarge}. These issues underscore the need for specialized strategies that address reliability, bias, and context sensitivity in \llm outputs.



\llms training can be broadly categorized into two stages: pre-training, which generally relies on a next-token prediction objective over large-scale corpora, and post-training, encompassing multiple rounds of fine-tuning and alignment.
Post-training mechanisms aim to mitigate \llms limitations by refining model behavior and aligning outputs with human intent, mitigating biases or inaccuracies~\cite{bai2022constitutional}. 

Adapting \llms to domain-specific tasks often involves techniques like \textbf{fine-tuning}~\cite{blob-RFT,lobo2024impact,trung2024reft}, which enables task-specific learning but risks overfitting and incurs high computational costs. To address these challenges, approaches such as \textbf{Reinforcement Learning (\rl)}~\cite{rafailov2024direct,ouyang2022training,shao2024deepseekmath} enhance adaptability by leveraging dynamic feedback and optimizing sequential decision-making. Additionally, advances in \textbf{scaling} techniques, including Low-Rank Adaptation (\lora)~\cite{hu2021lora}, adapters~\cite{Ebrahimi2024CROMECA, tafjord2021proofwritergeneratingimplicationsproofs}, and Retrieval-Augmented Generation (\rag)~\cite{asai2023self,gao2023retrieval,hu2024rethinking}, improve both computational efficiency and factual accuracy. These strategies, coupled with distributed training frameworks, facilitate large-scale deployment and further boost the usability of \llms across diverse applications (Figure~\ref{fig:intro_main}). 
Through these targeted post-training interventions, \llms become better aligned with human intent and ethical requirements, ultimately enhancing their real-world applicability.
Below, we summarize key post-training stages.

\noindent\textbf{a) Fine-Tuning in \llms:}
Fine-tuning adapts pre-trained \llms to specific tasks or domains by updating parameters on curated datasets~\cite{yue2023disc, luo2023empirical, li2023better, blob-RFT, lobo2024impact, openai-RFT, trung2024reft}. While \llms generalize well after large-scale pretraining, fine-tuning enhances performance in tasks like sentiment analysis~\cite{zhang2023sentiment, maas2011learning}, question answering, and domain-specific applications such as medical diagnosis~\cite{wang2022recognizing, luo2022biogpt, wysocki2024llm}. This process, typically supervised, aligns models with task requirements but poses challenges like overfitting, high computational costs, and sensitivity to data biases~\cite{trung2024reft, deng2024efficient, liu2024deepseek}. To this end, parameter-efficient techniques like \lora~\cite{hu2021lora} and adapters learn task-specific adaptation by updating explicit parameters, significantly reducing computational overhead. As models specialize, they may struggle with out-of-domain generalization, underscoring the trade-off between specificity and versatility.

\insightbox{ Fine-tuning tailors \llms for specific tasks, improving performance but risking overfitting, high compute costs, and reduced generalization.}

\noindent\textbf{b) Reinforcement Learning in \llms:}
In conventional \rl, an agent interacts with a structured environment, taking discrete actions to transition between states while maximizing cumulative rewards~\cite{schulman2017proximal}. 
\rl domains—such as robotics, board games, and control systems—feature well-defined state-action spaces and clear objectives~\cite{xie2024monte, shanahan2023role}.  
\rl in \llms differs significantly. Instead of a finite action set, \llms select tokens from a vast vocabulary, and their evolving state comprises an ever-growing text sequence~\cite{liu2024deepseek, shao2024deepseekmath, xu2024perfect, rafailov2024direct}. This complicates planning and credit assignment, as the impact of token selection may only emerge later. Feedback in language-based \rl is also sparse \cite{cao2024sparserewardsenhancingreinforcement}, subjective, and delayed, relying on heuristic evaluations and user preferences rather than clear performance metrics~\cite{dubois2024alpacafarm, shani2024multi, ouyang2022training}.  
Additionally, \llms must balance multiple, sometimes conflicting, objectives, unlike conventional \rl, which typically optimizes for a single goal. Hybrid approaches combining process-based rewards (e.g., chain-of-thought reasoning) with outcome-based evaluations (e.g., response quality) help refine learning~\cite{wei2022chain, li2024falcon, nakano2021webgpt}. Thus, \rl for \llms requires specialized optimization techniques to handle high-dimensional outputs, non-stationary objectives, and complex reward structures, ensuring responses remain contextually relevant and aligned with user expectations.

\insightbox{Reinforcement in \llms extends beyond conventional \rl as it navigates vast action spaces, handles subjective and delayed rewards, and balances multiple objectives, necessitating specialized optimization techniques.}

\noindent\textbf{c) Test Time Scaling in \llms:}
Test Time Scaling is optimizing model performance and efficiency without altering the core architecture. It enables better generalization while minimizing computational overhead. It is crucial for enhancing the performance and efficiency of \llms. It helps improve generalization across tasks but introduces significant computational challenges~\cite{gao2023scaling, snell2024scaling}. Balancing performance and resource efficiency requires targeted strategies at inference. Techniques like \texttt{CoT}~\cite{wei2022chain} reasoning and Tree-of-Thought (\texttt{ToT})~\cite{yao2024tree} frameworks enhance multi-step reasoning by breaking down complex problems into sequential or tree-structured steps. Additionally, search-based techniques\cite{jiang2014searching, xie2024selfbeambearchreasoning, tian2024toward, gandhi2024stream} enable iterative exploration of possible outputs, helping refine responses and ensure higher factual accuracy. These approaches, combined with methods like \lora~\cite{hu2021lora}, adapters, and  \rag~\cite{asai2023self,gao2023retrieval, yang2023leandojotheoremprovingretrievalaugmented}, optimize the model’s ability to handle complex, domain-specific tasks at scale.
 \rag enhances factual accuracy by dynamically retrieving external knowledge, mitigating limitations of static training data~\cite{gao2023retrieval, lewis2020retrieval, sun2024retrievalaugmentedhierarchicalincontextreinforcement}. Distributed training frameworks leverage parallel processing to manage the high computational demands of large-scale models.  
Test-time scaling optimizes inference by adjusting parameters dynamically based on task complexity~\cite{snell2024scaling, davis2024testing}. Modifying depth, width, or active layers balances computational efficiency and output quality, making it valuable in resource-limited or variable conditions.  
Despite advancements, scaling presents challenges such as diminishing returns, longer inference times, and environmental impact, especially when search techniques are performed at test time rather than during training~\cite{gao2023scaling}. Ensuring accessibility and feasibility is essential to maintain high-quality, efficient \llm deployment.
\insightbox{Test-time scaling enhances the adaptability of \llms by dynamically adjusting computational resources during inference. }

\subsection{Prior Surveys}
Recent surveys on \rl and \llms provide valuable insights but often focus on specific aspects, leaving key post-training components underexplored~\cite{ji2023survey, wang2024survey, wu2025survey, chang2024survey}. Many works examine \rl techniques like Reinforcement Learning from Human Feedback (\rlhf)~\cite{ouyang2022training}, Reinforcement Learning from AI Feedback (\texttt{RLAIF})~\cite{lee2023rlaif}, and Direct Preference Optimization (\dpo)~\cite{rafailov2024direct}, yet they overlook fine-tuning, scaling, and critical benchmarks essential for real-world applications. Furthermore, these studies have not explored the potential of \rl even without human annotation supervised finetuning in various frameworks such as DeepSeek R1 with \grpo \cite{shao2024deepseekmath}.  
Other surveys explore \llms in traditional \rl tasks, such as multi-task learning and decision-making, but they primarily classify \llm functionalities rather than addressing test-time scaling and integrated post-training strategies~\cite{dong2022survey, zhao2023survey}. Similarly, studies on \llm reasoning~\cite{han2024folionaturallanguagereasoning, xi2024selfpolishenhancereasoninglarge, saparov2023languagemodelsgreedyreasoners, lobo2024impact, liu2024improving, kojima2022large, deng2021reasonbert, sawada2023arbadvancedreasoningbenchmark} discuss learning-to-reason techniques but lack structured guidance on combining fine-tuning, \rl, and scaling. 
The absence of tutorials, along with reviews of software libraries and implementation tools, further limits their practicality. 
In contrast, this survey offers a comprehensive view of \llm post-training as shown in Figure~\ref{fig:intro_main} by systematically covering fine-tuning, \rl, and scaling as interconnected optimization strategies. We offer practical resources—benchmarks, datasets, and tutorials—to aid \llm refinement for real-world applications.

\subsection{Contributions}
The key contributions of this survey are as follows:

\begin{itemize}
    \item We provide a comprehensive and systematic review of post-training methodologies for \llms, covering \textbf{fine-tuning}, \textbf{\rl}, and \textbf{scaling} as integral components of model optimization.
    

    \item We offer a \textbf{structured taxonomy} of post-training techniques, clarifying their roles and interconnections, and present insights into open challenges and future research directions in optimizing \llms for real-world deployment.
    
    
    \item Our survey provides practical guidance by introducing key \textbf{benchmarks, datasets, and evaluation metrics} essential for assessing post-training effectiveness, ensuring a structured framework for real-world applications.
    
\end{itemize}

%% file: content/background.tex
\section{Background}\label{sec:background}
The \llms have transformed reasoning by learning to predict the next token in a sequence based on vast amounts of text data~\cite{radford2018improving, devlin2018bert} using Maximum Likelihood Estimation (\texttt{MLE}) \cite{myung2003tutorial, yang2019xlnet, shao2020bert}, which maximizes the probability of generating the correct sequence given an input. This is achieved by minimizing the negative log-likelihood:
\[
\mathcal{L}_{\text{MLE}} = -\sum_{t=1}^{T} \log P_{\theta}(y_t \mid y_{<t}, X).
\]
Here, \(X\) represents the input, such as a prompt or context. \(Y = (y_1, y_2, ..., y_T)\) is the corresponding target output sequence, and \(P_{\theta}(y_t \mid y_{<t}, X)\) denotes the model’s predicted probability for token \(y_t\), given preceding tokens.
\insightbox{Token-wise training can ensure fluency but may cause cascading errors due to uncorrected mistakes in inference.}
As these models scale, they exhibit emergent reasoning abilities, particularly when trained on diverse data that include code and mathematical content~\cite{wei2022emergent, wei2022chain}. However, despite their impressive capabilities, \llms struggle to maintain coherence and contextual relevance over long sequences. Addressing these limitations necessitates a structured approach to sequence generation, which naturally aligns with \rl. 

Since \llms generate text autoregressively—where each token prediction depends on previously generated tokens—this process can be modeled as a sequential decision-making problem within a Markov Decision Process (\mdp)~\cite{bellman1957markovian}. In this setting, the state \( s_t \) represents the sequence of tokens generated so far, the action \( a_t \) is the next token, and a reward \( R(s_t,a_t) \) evaluates the quality of the output. An \llm’s policy \(\pi_\theta\) is optimized to maximize the expected return:
\[
J(\pi_\theta) = \mathbb{E}\Bigl[\sum_{t=0}^\infty \gamma^t R(s_t,a_t)\Bigr],
\]
where \(\gamma\) is the discount factor that determines how strongly future rewards influence current decisions. A higher \(\gamma\) places greater importance on long-term rewards.
The primary objective in \rl is to learn a policy that maximizes the expected cumulative reward, often referred to as the return. This requires balancing exploration—trying new actions to discover their effects—and exploitation—leveraging known actions that yield high rewards. While \llms optimize a likelihood function using static data, \rl instead optimizes the expected return through dynamic interactions.
To ensure that \llms generate responses that are not only statistically likely but also aligned with human preferences, it is essential to go beyond static optimization methods. While likelihood-based training captures patterns from vast corpora, it lacks the adaptability needed for refining decision-making in interactive settings. By leveraging structured approaches to maximizing long-term objectives, models can dynamically adjust their strategies, balancing exploration and exploitation to improve reasoning, coherence, and alignment~\cite{hao2024llmreasonersnewevaluation, geiping2025scalingtesttimecomputelatent, li2025imaginereasoningspacemultimodal,ye2025limoreasoning}.
\insightbox{\llms exhibit emergent abilities due to scale, while \rl refines and aligns them for better reasoning and interaction.}

\subsection{\rl based Sequential Reasoning.}\label{SequentialReasoning}
The chain-of-thought reasoning employed in modern \llms is naturally framed as an \rl problem. In this perspective, each intermediate reasoning step is treated as an action contributing to a final answer. The objective function \( J(\pi_\theta) \) represents the expected reward of the policy \( \pi_\theta \), capturing how well the model performs over multiple reasoning steps. The policy gradient update is given by:
\[
\nabla_\theta J(\pi_\theta) = \mathbb{E}_\tau\Biggl[\sum_{t=1}^T \nabla_\theta \log \pi_\theta(x_t\mid x_{1:t-1})\, A(s_t,a_t)\Biggr],
\]
where the advantage function \(A(s_t,a_t)\) distributes credit to individual steps, ensuring that the overall reasoning process is refined through both immediate and delayed rewards. 
Such formulations, including step-wise reward decomposition \cite{lightman2023let,chen2024step}, have been crucial for enhancing the interpretability and performance of \llms on complex reasoning tasks.
In traditional \rl formulations, an agent has:
\[
\text{Value function: } V(s) \;=\; \mathbb{E}\bigl[ \text{future return} \,\mid\, s\bigr],
\]
\[
\text{Action-value (Q-) function: } Q(s, a) \;=\; \mathbb{E}\bigl[\text{future return}\,\mid\, s, a\bigr],
\]
\[
\text{Advantage function: } A(s, a) \;=\; Q(s, a) - V(s).
\]
In words, $A(s,a)$ measures \emph{how much better or worse} it is to take a specific action $a$ in state $s$ compared to what the agent would \emph{normally} expect (its baseline $V(s)$). 
\subsection{Early \rl Methods for Language Modeling.}
Here, we briefly overview pioneering methods that laid the groundwork for applying \rl to language generation tasks. These initial efforts train a decision-making model (policy ($p_\theta$)) by directly adjusting its parameters to maximize rewards. Some policy gradient approaches are explained below:

\noindent\textbf{Policy Gradient (\reinforce).}  
The \reinforce algorithm~\cite{nguyen2017reinforcement,williams1982learning} is a method used to improve decision-making by adjusting the model’s strategy (policy) based on rewards received from its actions. Instead of directly learning the best action for every situation, the algorithm refines how likely different actions are to be chosen, gradually improving outcomes over time.
At each step, the model updates its parameters (\(\theta\)) based on how well its past decisions performed:
\[
\theta \leftarrow \theta + \alpha \Bigl( G - b \Bigr) \sum_{t=1}^{T} \nabla_\theta \log \pi_\theta(a_t \mid s_t).
\]
Here: \( G \) represents the total reward the model accumulates over an episode, \( b \) is a baseline value that helps reduce variance, making learning more stable, \( \nabla_\theta \log \pi_\theta(a_t \mid s_t) \) measures how much a small change in \(\theta\) affects the probability of choosing action \(a_t\) given state \(s_t\), \( \alpha \) is the learning rate, controlling how much the policy updates at each step.

\insightbox{Optimizing actions based on long-term rewards, which account for the cumulative benefits of a sequence of reasoning steps rather than just immediate outcomes, is fundamental in recent LLMs. This approach allows models to explore multiple reasoning paths more effectively.}

\noindent\textbf{Curriculum Learning with \mixer.}. Ranzato et al. \cite{ ranzato2016sequenceleveltrainingrecurrent} introduces a gradual transition from maximum likelihood estimation (\mle) to \rl. The overall loss is a weighted combination:
\[
\mathcal{L} = \lambda(t) \, \mathcal{L}_{\text{\mle}} + \bigl(1-\lambda(t)\bigr) \, \mathcal{L}_{\text{\rl}},
\]
where \(\lambda(t)\) decreases with training time. This curriculum helps the model ease into the \rl objective and mitigate the mismatch between training and inference.

\noindent\textbf{Self-Critical Sequence Training (\scst).}  
\scst~\cite{rennie2017self} refines the policy gradient method by comparing the model’s sampled outputs against its own best (greedy) predictions. Instead of using an arbitrary baseline, \scst uses the model’s own highest-scoring output, ensuring that updates directly improve performance relative to what the model currently considers its best response.
The gradient update follows:
\[
\nabla_\theta J(\pi_\theta) \approx \Bigl( r(y^s) - r(\hat{y}) \Bigr) \nabla_\theta \log \pi_\theta(y^s),
\]
            where \( y^s \) is a sampled sequence, \( \hat{y} \) is the greedy output, and \( r(y) \) represents an evaluation metric such as \texttt{BLEU} \cite{papineni2002bleu} for translation or \texttt{CIDEr} \cite{vedantam2015ciderconsensusbasedimagedescription} for image captioning. Since the learning signal is based on the difference \( r(y^s) - r(\hat{y}) \), the model is explicitly trained to generate outputs that score higher than its own baseline under the evaluation metric. If the sampled output outperforms the greedy output, the model reinforces it; otherwise, it discourages that sequence. This direct feedback loop ensures that training aligns with the desired evaluation criteria rather than just maximizing likelihood. By leveraging the model’s own best predictions as a baseline, \scst effectively reduces variance and stabilizes training while optimizing real-world performance metrics.
\input{content/classification}

\noindent\textbf{Minimum Risk Training (\mrt).}  
\mrt~\cite{shen2016minimumrisktrainingneural} directly minimizes the expected risk over the output distribution. Given a task-specific loss \(\Delta(y, y^*)\) comparing the generated output \(y\) with the reference \(y^*\), the \mrt objective is defined as:
\[
\mathcal{L}_{\text{\mrt}}(\theta) = \sum_{y \in \mathcal{Y}} p_\theta(y \mid x) \, \Delta(y, y^*).
\]
This formulation incorporates evaluation metrics (e.g., \(1-\text{\texttt{BLEU}}\)) directly into training, enabling fine-grained adjustments of the policy.

\noindent\textbf{Advantage Actor-Critic (A2C/A3C).}  
\rl methods like \reinforce \cite{nguyen2017reinforcement} rely solely on policy gradients, which suffer from high variance, leading to unstable and inefficient learning. Since the reward signal fluctuates across different trajectories, updates may be noisy, causing slow or erratic convergence. To mitigate this, Actor-Critic methods~\cite{AC1999, bhatnagar2007incrementalac, a2c2016, a3c2016gpu} combine two components as follows: an actor and a critic. The actor is a policy \(\pi_\theta(a_t \mid s_t)\) that selects actions \(a_t\) at state \(s_t\), while the critic is a value function \(V_\phi(s_t)\) that evaluates the expected return of a state. The critic provides a more stable learning signal, reducing variance in policy updates and enabling efficient learning in continuous action spaces.
Actor updates are guided by the policy gradient theorem, where the advantage function 
$A(s_t, a_t) $ defined in Sec. \ref{SequentialReasoning}, 
determines how much better an action \(a_t\) is compared to the expected value of state \(s_t\). The policy with the learning rate $\alpha$ is updated as:
\[
\theta \leftarrow \theta + \alpha \, A(s_t, a_t) \, \nabla_\theta \log \pi_\theta(a_t \mid s_t).
\]
Meanwhile, the critic is updated using temporal difference learning, minimizing the squared error between its estimate and the actual return:
\[
\phi \leftarrow \phi - \beta \, \nabla_\phi \Bigl(V_\phi(s_t) - G_t\Bigr)^2.
\]
where $\beta$ is a learning rate for critic.
To enhance stability and efficiency, several improvements have been proposed. Eligibility traces allow learning from recent states, enabling faster convergence. Function approximation with neural networks ensures effective handling of high-dimensional inputs. Advanced variants such as Natural Gradient methods~\cite{kakade2001natural} adjust updates using the Fisher Information Matrix, improving convergence speed.  

A notable early example is Barto’s Actor-Critic model \cite{barto1983neuronlike}, where the critic uses a linear function \(V_\phi(s_t)\) and the actor follows a linear policy. Modern methods like A2C (Advantage Actor-Critic) \cite{a2c2016} and A3C (Asynchronous Advantage Actor-Critic) \cite{a3c2016gpu} extend this approach by parallelizing training across multiple environments, leading to faster and more stable learning.
By leveraging the critic’s value estimation, actor-critic methods stabilize learning, improve sample efficiency, and accelerate convergence, making them more effective for complex decision-making tasks.

\noindent\textbf{Connection with Modern Methods.}  
The aforementioned early \rl methods—\reinforce~\cite{nguyen2017reinforcement}, \mixer~\cite{ranzato2016sequenceleveltrainingrecurrent}, \seqgan~\cite{yu2017seqgansequencegenerativeadversarial}, \scst~\cite{rennie2017self}, \mrt~\cite{shen2016minimumrisktrainingneural}, and actor-critic algorithms established the mathematical foundations for sequential reasoning in \llms. These methods provided initial solutions to challenges such as exposure bias and high variance. Modern techniques such as large-scale \rl from Human Feedback (\rlhf) using \ppo \cite{schulman2017proximal} and advanced reward models, e.g., Group Relative Policy Optimization (\grpo) \cite{yang2024qwen2} build directly upon these ideas. By integrating sophisticated reward signals and leveraging efficient policy updates, contemporary \llms achieve improved reasoning, safety, and alignment with human values and pave the way for robust multi-step reasoning and improved quality of generated text. Table~\ref{tab:overview-RL-LLMs} provides an overview of recent models, including their parameters, architecture types, and the distilled \rl methods employed, along with links for easy access.

%% file: content/classification.tex
\begin{table*}[htb!]
\centering
\small
\begin{adjustbox}{max width=1.0\textwidth}
\begin{threeparttable}
\begin{tabular}{l>{\centering\arraybackslash}m{2.5cm}c>{\centering\arraybackslash}m{3cm}ccc>{\centering\arraybackslash}m{2.5cm}c>{\centering\arraybackslash}m{2.5cm}}
\toprule
\rowcolor{gray!8}
\textbf{RL Enhanced LLMs} & \textbf{Developer} & \textbf{Source} & \textbf{\# Params} & \textbf{RL Methods} & \textbf{Fine-Tuning} & \textbf{Architecture Type} & \textbf{Model} & \textbf{TTS} \\
\midrule

\deepseek \citep{liu2024deepseek} & Deepseek & \href{https://www.deepseek.com/}{Link} & 236B-A21B & GRPO & DPO + GRPO & MoE & Open & \textcolor{blue}{\checkmark} \\
\texttt{GPT 4.5} \citep{openai2025gpt45systemcard} & OpenAI & \href{https://openai.com/gpt-4}{Link} & - & RLHF, PPO, RBRM & SFT + RLHF & MoE & Closed & \textcolor{blue}{\checkmark} \\

\gemini \citep{team2023gemini} & Google & \href{https://gemini.google.com/app}{Link} & - & RLHF & SFT + RLHF & Single Model & Closed & \textcolor{red}{\ding{55}} \\

\texttt{Claude 3.7} \citep{anthropic2025claude37} & Anthropic & \href{https://www.anthropic.com/claude/sonnet}{Link} & - & RLAIF & SFT + RLAIF & Single Model & Closed & \textcolor{red}{\ding{55}} \\
\reka \citep{team2024reka} & Reka & \href{https://www.reka.ai/}{Link} & 7B, 21B & RLHF, PPO & SFT + RLHF & Single Model & Closed & \textcolor{red}{\ding{55}} \\

\texttt{DeepSeekR1} \citep{guo2025deepseek} & Deepseek & \href{https://www.deepseek.com/}{Link}  & 240B-A22B & GRPO & DPO + GRPO & MoE & Open & \textcolor{blue}{\checkmark} \\

\nemotron \citep{adler2024nemotron} & NVIDIA & \href{https://huggingface.co/nvidia/Nemotron-4-340B-Instruct}{Link} & 340B & DPO, RPO & DPO + RPO & Single Model & Closed & \textcolor{red}{\ding{55}} \\
\texttt{Falcon} \citep{almazrouei2023falconseriesopenlanguage} & TII & \href{https://huggingface.co/tiiuae/falcon}{Link} & 40B & - & SFT & Single Model & Open & \textcolor{red}{\ding{55}} \\
\gptfour \citep{openai2023gpt} & OpenAI & \href{https://openai.com/gpt-4}{Link} & - & RLHF, PPO, RBRM & SFT + RLHF & MoE & Closed & \textcolor{blue}{\checkmark} \\
\llama \citep{dubey2024llama} & Meta & \href{https://github.com/facebookresearch/llama}{Link} & 8B, 70B, 405B & DPO & SFT + DPO & Single Model & Open & \textcolor{red}{\ding{55}} \\

\qwen \citep{hui2024qwen2} & Alibaba & \href{https://huggingface.co/Qwen}{Link} & (0.5-72)B, 57B-A14B & DPO & SFT + DPO & Single Model & Open & \textcolor{blue}{\checkmark} \\

\gemma \citep{team2024gemma} & Google & \href{https://ai.google.dev/gemma}{Link} & 2B, 9B, 27B & RLHF & SFT + RLHF & Single Model & Open & \textcolor{red}{\ding{55}} \\
\starling \citep{zhu2024starling} & Berkeley & \href{https://huggingface.co/berkeley-nest/Starling-LM-7B-alpha}{Link} & 7B & RLAIF, PPO & SFT + RLAIF & Single Model & Open & \textcolor{red}{\ding{55}} \\
\texttt{Moshi} \citep{kyutai2024moshi} & Kyutai & \href{https://github.com/kyutai-labs/moshi}{Link} & 7B & - & - & Multi-modal & Open & \textcolor{blue}{\checkmark} \\
\athene \citep{nexusflow2024athene} & Nexusflow & \href{https://nexusflow.ai/blogs/athene}{Link} & 70B & RLHF & SFT + RLHF & Single Model & Open & \textcolor{red}{\ding{55}} \\
\texttt{GPT-3.5} \citep{openai2023gpt} & OpenAI & \href{https://chatgpt.com/g/g-F00faAwkE-open-a-i-gpt-3-5}{Link} & 3.5B, 175B & RLHF, PPO & SFT + RLHF & MoE & Closed & \textcolor{blue}{\checkmark} \\
\hermes \citep{teknium2024hermes} & Nous & \href{https://huggingface.co/NousResearch/Nous-Hermes-13b}{Link} & 8B, 70B, 405B & DPO & SFT + DPO & Single Model & Open & \textcolor{red}{\ding{55}} \\

\texttt{Zed} \citep{team2025zed} & Zed AI & \href{https://zed.dev/ai}{Link} & 500B & RLHF & RLHF & Multi-modal & Open & \textcolor{blue}{\checkmark} \\
\texttt{PaLM 2}~\cite{anil2023palm}  & Google & \href{https://blog.google/technology/ai/google-palm-2-ai-large-language-model/}{Link} & - & RLHF & - & Single Model & Closed & \textcolor{blue}{\checkmark} \\
\internlm \citep{cai2024internlm2} & SAIL & \href{https://chat.intern-ai.org.cn/}{Link} & 1.8B, 7B, 20B & RLHF, PPO & SFT + RLHF & Single Model & Closed & \textcolor{red}{\ding{55}} \\

\texttt{Supernova} \citep{supernova2025} & Nova AI & \href{https://huggingface.co/theNovaAI/Supernova-experimental}{Link} & 220B & RLHF & RLHF & Multi-modal & Open & \textcolor{blue}{\checkmark} \\
\texttt{Grok3} \citep{xai2025grok3} & Grok-3 & \href{https://x.ai/}{Link} & 175B & - & DPO & Dense & Open & \textcolor{blue}{\checkmark} \\

\texttt{Pixtral} \citep{agrawal2024pixtral12b} & Mistral AI & \href{https://mistral.ai/}{Link} & 12B, 123B & - & PEFT & Multimodal & Open  & \textcolor{blue}{\checkmark} \\

\texttt{Minimaxtext} \citep{minimax2025minimax01scalingfoundationmodels} & MiniMax & \href{https://www.minimax.ai/}{Link} & 456B & - & SFT & Single Model & Closed & \textcolor{red}{\ding{55}} \\

\texttt{Amazonnova} \citep{Intelligence2024} & Amazon & \href{https://aws.amazon.com/bedrock/}{Link} & - & DPO, RLHF, RLAIF & SFT & Single Model & Closed & \textcolor{red}{\ding{55}} \\

\texttt{Fugakullm} \citep{fujitsu2024fugaku} & Fujitsu & \href{https://www.fujitsu.com/global/about/resources/news/press-releases/2024/fugaku-llm.html}{Link} & 13B & - & - & Single Model & Closed & \textcolor{red}{\ding{55}} \\
\texttt{Nova} \citep{rubiks2024nova} & Rubik's AI & \href{https://rubiks.ai/}{Link} & - & - & SFT & Proprietary & Closed & \textcolor{red}{\ding{55}} \\
\texttt{03}~\cite{o3} & OpenAI & \href{https://openai.com/index/openai-o3-mini/}{Link} & - & RL through CoT & RL through CoT & Single Model & Closed & \textcolor{blue}{\checkmark} \\
\texttt{Dbrx} \citep{mosaic2024introducing} & Databricks & \href{https://www.databricks.com/blog/dbrx}{Link} & 136B & - & SFT & Single Model & Open & \textcolor{red}{\ding{55}} \\
\instructgpt~\citep{ouyang2022training} & OpenAI & \href{https://gpt3demo.com/apps/instructgpt}{Link} & 1.3B, 6B, 175B & RLHF, PPO & SFT + RLHF & Single Model & Closed & \textcolor{red}{\ding{55}} \\
\texttt{Openassistant} \citep{köpf2023openassistantconversationsdemocratizing} & LAION & \href{https://open-assistant.io/}{Link} & 17B & - & SFT & Single Model & Open & \textcolor{red}{\ding{55}} \\
\chatglm \citep{glm2024chatglm} & Zhipu AI & \href{https://github.com/THUDM/ChatGLM-6B}{Link} & 6B, 9B & ChatGLM-RLHF & SFT + RLHF & Single Model & Open & \textcolor{red}{\ding{55}} \\
\zephyr \citep{zephyr_141b} & Argilla & \href{https://huggingface.co/blog/Isamu136/understanding-zephyr}{Link} & 141B-A39B & ORPO & DPO + ORPO & MoE & Open & \textcolor{blue}{\checkmark} \\
\texttt{phi-3} \citep{abdin2024phi} & Microsoft & 
\href{http://news.microsoft.com/source/features/ai/the-phi-3-small-language-models-with-big-potential/}{Link} & 3.8B, 7B, 14B & DPO & SFT + DPO & Single Model & Closed & \textcolor{red}{\ding{55}} \\
\texttt{Jurassic} \citep{lieber2021jurassic} & AI21 Labs & \href{https://www.ai21.com/}{Link} & - & - & SFT & Proprietary & Closed & \textcolor{red}{\ding{55}} \\
\texttt{Kimi K1.5} \citep{team2025kimi} & Moonshot AI & \href{https://github.com/MoonshotAI/Kimi-k1.5}{Link} & 150B & - & RLHF & Multi-modal & Open & \textcolor{blue}{\checkmark} \\

\texttt{Phi-4} \citep{abdin2024phi4technicalreport} & Microsoft & \href{https://ollama.com/library/phi4}{Link} & 28B, 70B, 140B & DPO & SFT + DPO & Single Model & Closed & \textcolor{red}{\ding{55}} \\
\texttt{Chameleon} \citep{team2024chameleon} & Meta AI & \href{https://ai.meta.com/blog/chameleon-meta-ai/}{Link} & 34B & - & SFT & Single Model & Open & \textcolor{red}{\ding{55}} \\

\texttt{Cerebrasgpt} \citep{dey2023cerebrasgptopencomputeoptimallanguage} & Cerebras & \href{https://huggingface.co/cerebras/Cerebras-GPT}{Link} & 13B & - & SFT & Single Model & Open & \textcolor{red}{\ding{55}} \\

\texttt{Bloomberggpt} \citep{wu2023bloomberggptlargelanguagemodel} & Bloomberg L.P. & \href{https://www.bloomberg.com/company/press/bloombergGPT}{Link} & 50B & - & SFT & Single Model & Closed & \textcolor{red}{\ding{55}} \\

\texttt{Chinchilla} \citep{hoffmann2022trainingcomputeoptimallargelanguage} & DeepMind & \href{https://www.deepmind.com/publications/training-a-high-performance-language-model-using-chinchilla}{Link} & 70B & RLHF, PPO & SFT & Single Model & Closed & \textcolor{red}{\ding{55}} \\



\bottomrule
\end{tabular}
\end{threeparttable}
\end{adjustbox}

\caption{An overview of reinforcement learning-enhanced \llms, where '141B-A39B' denotes a Mixture of Experts (\texttt{MoE}) model with 141 billion total parameters, of which 39 billion are utilized during inference. TTS stands for Test-Time Scaling.}
\label{tab:overview-RL-LLMs}
\end{table*}

%% file: content/methods.tex
\section{Reinforced \llms}
\label{sec_reinforcce}
From a methodological perspective, the integration of \rl into \llm reasoning typically follows three core steps:
\begin{enumerate}
    \item \textbf{Supervised Fine-Tuning (\texttt{SFT})}: Commences with a pretrained language model that is subsequently refined on a supervised dataset of high-quality, human-crafted examples. This phase ensures the model acquires a baseline compliance with format and style guidelines.
    \item \textbf{Reward Model (\texttt{RM}) Training}: Generated outputs from the fine-tuned model are collected and subjected to human preference labeling. The reward model is then trained to replicate these label-based scores or rankings, effectively learning a continuous reward function that maps response text to a scalar value.
    \item \textbf{\rl Fine-Tuning}: Finally, the main language model is optimized via a policy gradient algorithm most e.g \ppo to maximize the reward model's output. By iterating this loop, the \llm learns to produce responses that humans find preferable along key dimensions such as accuracy, helpfulness, and stylistic coherence.
    \item \textbf{Reward Modeling and Alignment:} Sophisticated reward functions are developed—drawing from human preferences, adversarial feedback, or automated metrics—to guide the model toward outputs that are coherent, safe, and contextually appropriate. These rewards are critical for effective credit assignment across multi-step reasoning processes.
\end{enumerate}
Early approaches to aligning \llms with human preferences leveraged classical \rl algorithms, such as  \ppo~\cite{schulman2017proximal} and Trust Region Policy Optimization (\trpo)~\cite{schulman2017trustregionpolicyoptimization}, which optimize a policy by maximizing the expected cumulative reward while enforcing constraints on policy updates via a surrogate objective function and KL-divergence regularization~\cite{vieillard2020leverage}. 
Improved alternatives to these methods for scalable preference-based optimization have emerged, such as Direct Preference Optimization (\dpo)~\cite{rafailov2024direct,guo2024direct} and Group Relative Policy Optimization (\grpo)~\cite{yang2024qwen2, shao2024deepseekmath, liu2024deepseek}, which reformulate the alignment objective as a ranking-based contrastive loss function \cite{an2022colo} over human-labeled preference data. 
Unlike \ppo and \trpo~\cite{schulman2017trustregionpolicyoptimization}, which rely on explicit reward models and critic networks, \dpo and \grpo directly optimize the policy by leveraging log-likelihood ratios and group-wise reward comparisons, respectively, eliminating the need for explicit value function approximation while preserving preference-consistent learning dynamics.
This transition from classical \rl-based alignment to preference-based direct optimization introduces novel formulations such as contrastive ranking loss, policy likelihood ratio regularization, and grouped advantage estimation, which are explained in subsequent sections.

\subsection{Reward modeling}
Let $\mathcal{X}$ be the space of possible \emph{queries} (e.g., user prompts). For each query $x \in \mathcal{X}$, we collect one or more \emph{candidate responses} $\{ y_j \}_{j=1}^{m_x}$
where $m_x$ is the number of candidate responses for query $x$. Typically, these responses are generated by a language model or policy under different sampling or prompting conditions.
Human annotators provide preference judgments for these responses. These can take various forms:
\begin{itemize}
    \item \textbf{Pairwise preference}: For two responses $y_j$ and $y_k$ to the same query $x$, an annotator indicates whether $y_j$ is preferred to $y_k$.
    \item \textbf{Rankings}: A partial or total ordering of the candidate responses, e.g.\ $y_{j_1} \succ y_{j_2} \succ \cdots \succ y_{j_{m_x}}$.
\end{itemize}
We denote such human preference data by $\{ r_j \}$ for each response or pair, where $r_j$ might be a label, a rank, or an index indicating preference level. The overall dataset $\mathcal{D}$ then consists of $N$ annotated examples:
\[
\mathcal{D} \;=\; \Big\{ (x^i, \{y_j^i\}_{j=1}^{m_i}, \{\text{preferences}^i\})\Big\}_{i=1}^N.
\]
In practice, a large number of queries $x$ are sampled from real or simulated user requests. Candidate responses $\{ y_j \}_{j=1}^{m_x}$ are generated by either sampling from a base language model or using beam search or other decoding strategies.
Human annotators then provide pairwise or ranking feedback on which responses are better (or worse) according to predefined criteria (e.g., quality, correctness, helpfulness, etc).
We train a parametric model Reward Model ($R_\theta(x, y)$), referred to as the \emph{reward model}, to map each (query, response) pair $(x, y)$ to a scalar score. The goal is for $R_\theta$ to reflect the \emph{alignment} or \emph{preference} level, such that:
\[
R_\theta : \mathcal{X} \times \mathcal{Y} \;\to\; \mathbb{R}.
\]
Here $\mathcal{Y}$ is the space of all possible responses.

To train $R_\theta$, we use the human preference labels in $\mathcal{D}$ to define a suitable \emph{ranking-based} loss, as explained below.

\textbf{I. Bradley--Terry Model (Pairwise).}
For pairwise preferences, Bradley-Terry model~\cite{bradley1952rank} is often used. Suppose the dataset indicates that, for a given query $x$, human annotators prefer $y_j$ to $y_k$, we denote it as $y_j \succ y_k$.
Under Bradley--Terry, the probability of $y_j$ being preferred over $y_k$ is given by:
\[
P\bigl(y_j \succ y_k \,\mid\, x;\theta \bigr)
\;=\;
\frac{\exp\bigl(R_\theta(x,\,y_j)\bigr)}{\exp\bigl(R_\theta(x,\,y_j)\bigr) \;+\; \exp\bigl(R_\theta(x,\,y_k)\bigr)}.
\]
We train $R_\theta$ by \emph{maximizing} the likelihood of observed preferences (or equivalently \emph{minimizing} the negative log-likelihood):
\begin{equation*}
\small
\mathcal{L}_{\text{BT}}(\theta) 
= -\!\! \sum_{\substack{(x,\,y_j \succ y_k) \,\in\, \mathcal{D}}}
\log P\bigl(y_j \succ y_k \mid x; \theta \bigr).
\end{equation*}
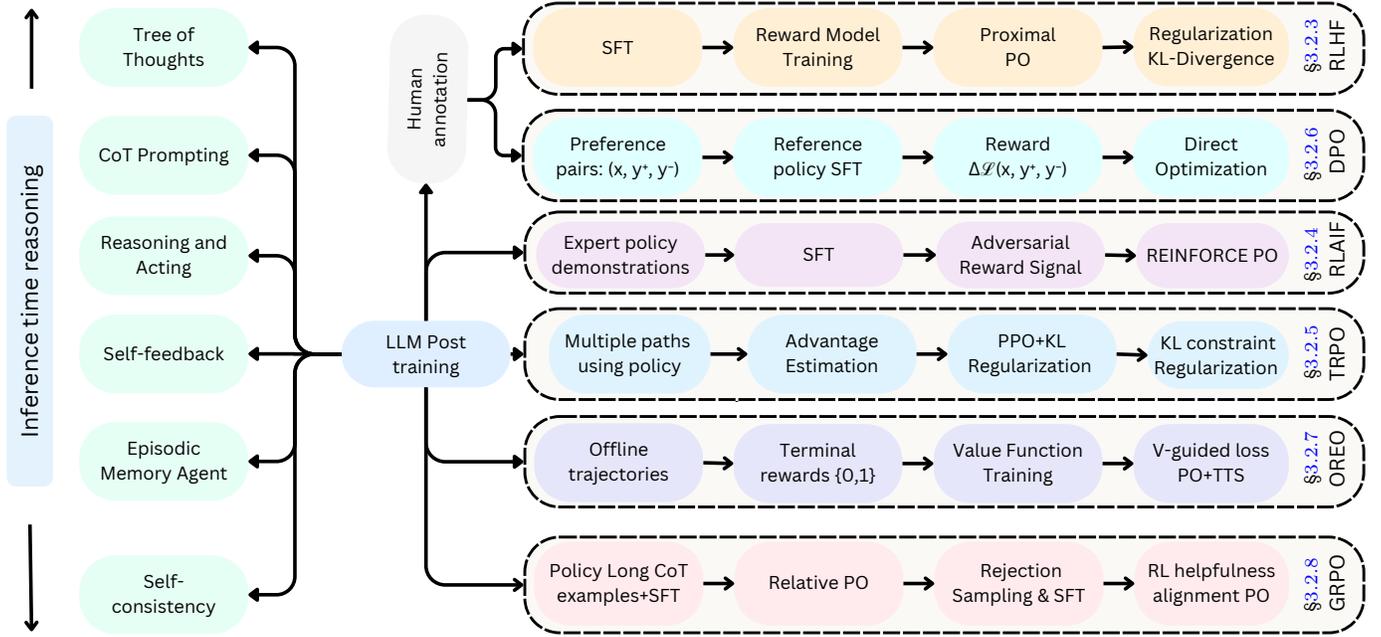
\begin{figure*}
    \centering
    \includegraphics[width=\linewidth]{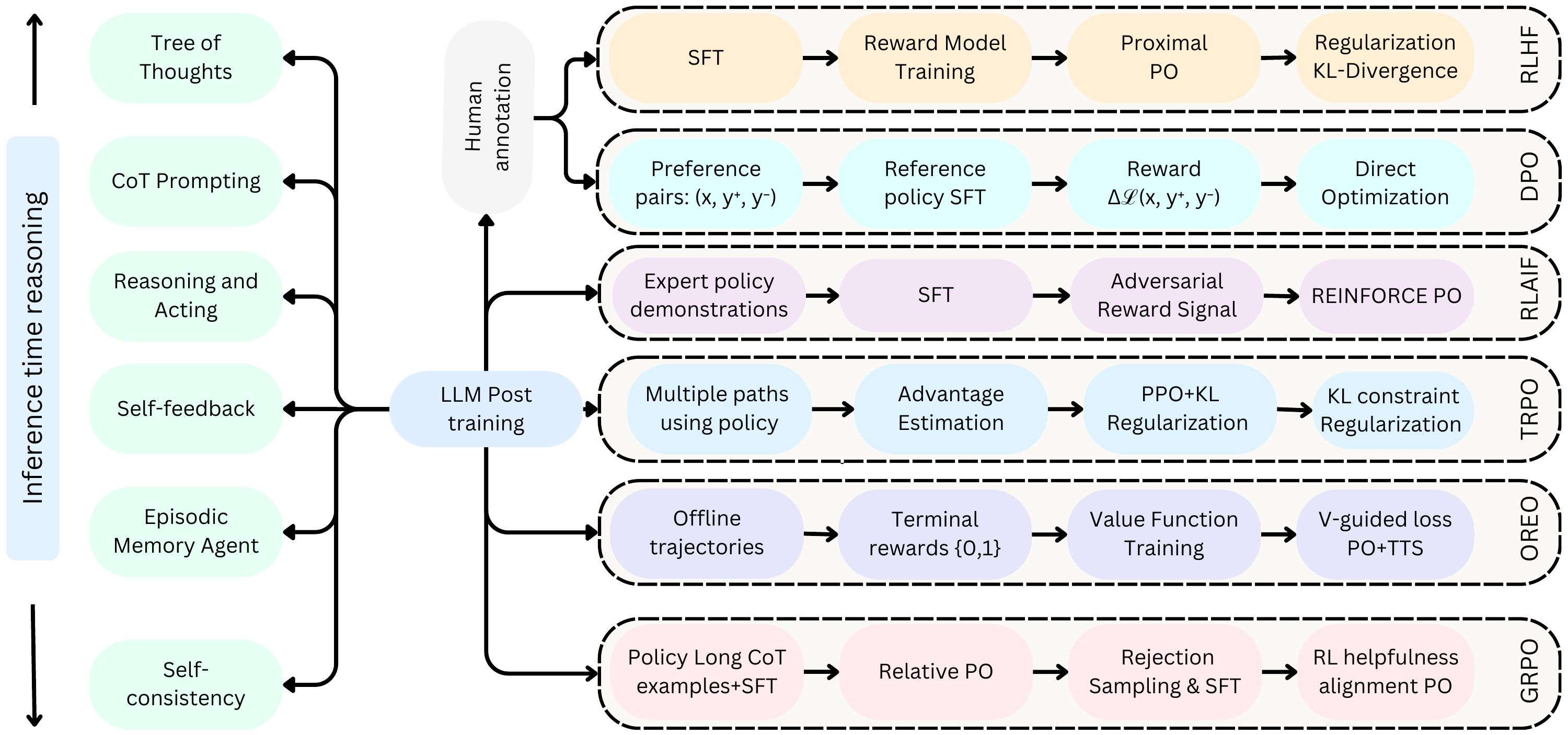}
    \put(-25,10){\rotatebox{90}{\scriptsize{\S \ref{sec_grpo}}}}
    \put(-25,58){\rotatebox{90}{\scriptsize{\S \ref{sec_oreo}}}} 
    \put(-25,98){\rotatebox{90}{\scriptsize{\S \ref{sub_trpo}}}} 
    \put(-25,135){\rotatebox{90}{\scriptsize{\S \ref{sub_rlaif}}}} 
    \put(-25,173){\rotatebox{90}{\scriptsize{\S \ref{sec_dpo}}}} 
    \put(-25,214){\rotatebox{90}{\scriptsize{\S \ref{sub_rlhf}}}} 
    \caption{Overview of Large Language Models (\llms) reasoning methods, showcasing pathways for enhancing reasoning capabilities through approaches like Chain-of-Thought (\cot) prompting, self-feedback, and episodic memory. The diagram highlights multiple reinforcement learning-based optimization techniques, including \grpo, \rlhf, \dpo, and \rlaif, for fine-tuning reasoning models with reward mechanisms and preference-based learning.}
    \label{fig:RL_overview}
\end{figure*}

\textbf{II. Plackett--Luce Model\footnote{\url{https://hturner.github.io/PlackettLuce/}} (Rankings).}
When full or partial \emph{rankings} of $m$ responses are available, i.e.,
\[
y_{j_1} \succ y_{j_2} \succ \cdots \succ y_{j_m},
\]
the Plackett--Luce model~\cite{plackett1975analysis} factorizes the probability of this ranking as:
\[
P\bigl(y_{j_1}, \ldots, y_{j_m} \,\mid\, x;\theta \bigr)
\;=\;
\prod_{\ell=1}^{m}
\frac{\exp\bigl(R_\theta(x, y_{j_\ell})\bigr)}
     {\sum_{k=\ell}^{m}
        \exp\bigl(R_\theta(x, y_{j_k})\bigr)}.
\]
Its negative log-likelihood is:
\[
\mathcal{L}_{\text{PL}}(\theta)
\;=\;
- \sum_{(x,\, \text{rank}) \in \mathcal{D}}
\sum_{\ell=1}^{m}
\log\!
\left(
   \frac{\exp\bigl(R_\theta(x, y_{j_\ell})\bigr)}
        {\sum_{k=\ell}^{m}
            \exp\bigl(R_\theta(x, y_{j_k})\bigr)}
\right).
\]
In practice, one minimizes the sum (or average) of the chosen ranking-based loss over all preference data:
\[
\mathcal{L}(\theta)
\;=\;
\frac{1}{|\mathcal{D}|}
\sum_{\substack{(x,\,\{y_j\},\,\text{prefs})\,\in\,\mathcal{D}}}
\mathcal{L}_{\text{ranking}}\Bigl(\theta; x, \{y_j\}, \text{prefs}\Bigr),
\]
where $\mathcal{L}_{\text{ranking}}$ could be either $\mathcal{L}_{\text{BT}}$ or $\mathcal{L}_{\text{PL}}$.
While the reward model $R_\theta(x,y)$ provides a \emph{scalar reward signal} reflecting human preferences,  this connects to common \rl concepts, especially the \emph{advantage function}. 
\insightbox{Reward modeling uses ranking-based losses to learn a function from human preferences for policy optimization.}
\noindent\textbf{Reward modeling Types.}  Rewards can be categorized into explicit and implicit approaches.
\subsubsection{Explicit Reward Modeling} 
Explicit reward modeling defines reward functions directly based on predefined rules, heuristics, or human annotations. This reward structure involves direct, numeric signals from humans or from specialized AI modules trained to approximate human judgments (e.g., ranking or pairwise comparison). 
This method can produce precise reward estimates but may be time-consuming or costly at scale. Illustrative use cases include `red-teaming' exercises where experts rate the severity of toxic outputs, or domain-specialist tasks in which correctness must be validated by a subject matter expert.

\subsubsection{Implicit Reward Modeling}  Implicit reward modeling infers rewards indirectly from observed behaviors, interactions, or preference signals, often leveraging machine learning techniques to uncover latent reward structures. It derives its signals from user interaction metrics such as upvotes, acceptance rates, click-through patterns, or session engagement times. While it can accumulate vast datasets with minimal overhead, this approach risks fostering behaviors that exploit engagement heuristics at the expense of content quality or veracity.


\noindent\textbf{Reward Function.} Defining a reward function for text generation tasks is an ill-posed problem \cite{rewardingprogress, lambert2024rewardbench}.
The existing \rl methods in \llms either focus on the generation process  outcome (Outcome Reward Modeling) or the (Process Reward Modeling), to shape \llm behaviors. We explain these two reward modeling paradigms below.

\subsubsection{Outcome Reward Modeling} 
Measures the end result (e.g., whether the final answer is factually correct or solves the user's query). This model is straightforward to implement but may offer limited insight into how the conclusion was reached. It is prevalent in short-response tasks, where the user's primary concern is the correctness or succinctness of the final statement. For long-response tasks, outcome based reward can lead to \emph{credit assignment problem}, i.e., which specific actions or states lead to a particular reward outcome.

\subsubsection{Process Reward Modeling} Assigns feedback at intermediate reasoning steps, incentivizing coherent, logically consistent, and well-structured chains of thought. This approach is particularly valuable for tasks involving mathematical derivations, legal arguments, or code debugging, in which the path to the answer is as significant as the final statement. 
In such problems, the reward assigned in individual steps encourages transparency and robust step-by-step reasoning. However, it requires a more complex annotation process, e.g., requires ``gold'' reasoning steps or partial credit scoring.
Process rewards can be combined with outcome rewards for a strong multi-phase training signal.
\insightbox{\texttt{Policy Reward Modeling} (\texttt{PRM}) with last-step aggregation outperforms \texttt{Outcome Reward Modeling} (\texttt{ORM}) by leveraging final-step evaluations to optimize policy updates more effectively.}

\subsubsection{Iterative \rl with Adaptive Reward Models}
Adaptive Reward Models is a training methodology designed to continuously improve the performance of \llms by iteratively refining the reward models and the policy model. This approach addresses the challenges of reward hacking and reward model drift, which can occur when the reward model becomes misaligned with the desired objectives during large-scale \rl training. The \rl process is divided into multiple iterations, where the model is trained in cycles. After each iteration, the reward model is updated based on the latest model behavior and human feedback. The reward model is not static but evolves over time to better align with human preferences and task requirements. This adaptation ensures that the reward signals remain accurate and relevant as the model improves. 
Repeat the iterative process until the model's performance plateaus or meets the desired benchmarks. The reward model and policy model co-evolve, with each iteration bringing them closer to optimal alignment.
\subsection{Policy Optimization}
\label{subsec:policy-optimization}
Once we have a trained reward model $R_\theta(x,y)$ that captures human preferences, we can integrate it into a \rl framework to \emph{optimize} a policy $\pi_\phi$. In essence, we replace (or augment) the environment’s native reward signal with $R_\theta(x,y)$ so that the agent focuses on producing responses $y$ that humans prefer for a given query $x$. 

In typical \rl notation:
\begin{itemize}
    \item Each \emph{state} $s$ here can be interpreted as the partial dialogue or partial generation process for the next token (in language modeling).
    \item Each \emph{action} $a$ is the next token (or next chunk of text) to be generated.
    \item The \emph{policy} $\pi_\phi(a\mid s)$ is a conditional distribution over the next token, parameterized by $\phi$.
\end{itemize}
We seek to find $\phi$ that \emph{maximizes} the expected reward under $R_\theta$. Concretely, let $x$ be a user query, and let $y \sim \pi_\phi(\cdot \mid x)$ be the generated response. We aim to solve:
\[
\max_{\phi} \; \mathbb{E}_{x \sim \mathcal{X}}\Bigl[\,
\mathbb{E}_{y \sim \pi_\phi(\cdot \,\mid\, x)}\bigl[R_\theta(x,y)\bigr]
\Bigr].
\]
This means that \emph{on average}, over user queries $x$ and responses $y$ drawn from the policy $\pi_\phi$, we want the reward model’s score $R_\theta(x,y)$ to be as high as possible.

\noindent\textbf{Policy Gradient and Advantage.}
The modern algorithms (e.g., PPO~\cite{schulman2017proximal}, \texttt{GRPO}~\cite{shao2024deepseekmath}, \trpo~\cite{schulman2017trustregionpolicyoptimization}) rely on \emph{policy gradients}. Figure \ref{fig:tts} presents a structured comparison of the these main \rl frameworks. Each framework builds upon different principles for policy learning, reference modeling, and reward computation. Recall that the \emph{advantage function} $A(s,a)$ quantifies how much better an action $a$ is than the baseline expected return $V(s)$. At a high level, we update the policy $\pi_\phi$ in the direction that increases $\pi_\phi(a \mid s)$ for actions $a$ with \emph{positive advantage} and decreases it for negative-advantage actions. Formally, the advantage $A_t$ at time $t$ can be written as:
\[
A_t = Q(s_t, a_t) 
\;-\; 
V(s_t),
\]
where $Q(s_t, a_t)$ is the expected future return (sum of future rewards, including $R_\theta$) starting from $s_t$ when taking action $a_t$.

When using the reward model $R_\theta$:
\begin{enumerate}
    \item We interpret $R_\theta(x,y)$ as the \emph{immediate} or \emph{terminal} reward for the generated response $y$.
    \item The policy’s \emph{future returns} thus factor in how likely subsequent tokens are to be positively scored by $R_\theta$.
    \item The \emph{advantage} function still captures how much better a particular generation step is compared to the baseline performance $V(s_t)$.
\end{enumerate}

\insightbox{The reward model learns relative preferences rather than absolute scores. This avoids the need for calibrated human ratings and focuses on pairwise comparisons.}

\subsubsection{Odds Ratio Preference Optimization (\texttt{ORPO})}
\input{figures/rhlf_flow_fig}
The simplest method is \texttt{ORPO}~\cite{hong2024orpomonolithicpreferenceoptimization} which \emph{directly} optimizing a policy from pairwise human preferences. Instead of first learning a separate reward model and then running standard \rl, \texttt{ORPO} updates the policy to increase the likelihood of \emph{preferred} responses (according to human labels) relative to \emph{dispreferred} ones. The key idea is to look at the \emph{odds ratio}:
\[
\frac{\pi_\phi(y_j \mid x)}
     {\pi_\phi(y_k \mid x)},
\]
where $y_j$ is the \emph{preferred} response and $y_k$ is the \emph{less-preferred} response for a given query $x$.

\textbf{Pairwise Preference Probability.}
In many direct preference approaches (e.g., Bradley--Terry style), one writes
\[
P_\phi\bigl(y_j \succ y_k \,\mid\, x\bigr)
\;=\;
\sigma\!\Bigl(\ln \frac{\pi_\phi(y_j \mid x)}{\pi_\phi(y_k \mid x)}\Bigr)
\;=\;
\frac{1}{1 + \exp\!\Bigl( \ln \frac{\pi_\phi(y_k \mid x)}{\pi_\phi(y_j \mid x)} \Bigr)},
\]
where $\sigma(\cdot)$ is the logistic (sigmoid) function. Intuitively, if the policy $\pi_\phi$ assigns higher probability to $y_j$ than to $y_k$, the \emph{odds} $\tfrac{\pi_\phi(y_j \mid x)}{\pi_\phi(y_k \mid x)}$ exceed 1, making $y_j$ more likely to be the \emph{preferred} outcome under the model.

In \texttt{ORPO}, one typically defines a \emph{negative log-likelihood} loss for all pairs $\{(x, y_j \succ y_k)\}$ in the dataset:
\[
\mathcal{L}_{\text{ORPO}}(\phi)
\;=\;
- \sum_{(x,\,y_j \succ y_k)\,\in\,\mathcal{D}}
\log 
\Bigl(
P_\phi\bigl(y_j \succ y_k \,\mid\, x\bigr)
\Bigr).
\]
Substituting the logistic form gives:
\[
\mathcal{L}_{\text{ORPO}}(\phi)
\;=\;
- \sum_{(x,\,y_j \succ y_k)\,\in\,\mathcal{D}}
\log
\Bigl(
\frac{ \pi_\phi(y_j \mid x) }
     { \pi_\phi(y_j \mid x) \;+\; \pi_\phi(y_k \mid x) }
\Bigr),
\]
which can also be interpreted as maximizing the \emph{log odds ratio} for the correct (preferred) label in each pairwise comparison.

\textbf{Interpretation via Odds Ratios.}
By treating each preference label ($y_j \succ y_k$) as a constraint on the \emph{odds} $\tfrac{\pi_\phi(y_j\mid x)}{\pi_\phi(y_k \mid x)}$, \texttt{ORPO} pushes the policy to increase its probability mass on $y_j$ while decreasing it on $y_k$. When viewed in logarithmic space:
\[
\ln \Bigl(\tfrac{\pi_\phi(y_j\mid x)}{\pi_\phi(y_k\mid x)}\Bigr),
\]
a higher value corresponds to a greater likelihood of selecting $y_j$ over $y_k$. Hence, minimizing $\mathcal{L}_{\text{ORPO}}(\phi)$ aligns $\pi_\phi$ with the human-labeled preferences.

\warningbox{ Odds Ratio Preference Optimization (\texttt{ORPO}) is potentially less flexible for combining multiple reward signals.}

\subsubsection{Proximal Policy Optimization (\ppo) in \llms}
\label{ppo_sub}
A popular method for policy optimization is \ppo~\cite{schulman2017proximal}, a strategy adapted to align \llms with human feedback. Given a policy $\pi_\theta$ parameterized by $\theta$ and a reward function $R$, \ppo updates the policy by optimizing a clipped objective that balances exploration and stability. Specifically, if $r_t(\theta) = \frac{\pi_\theta(a_t|s_t)}{\pi_{\theta_{\text{ref}}}(a_t|s_t)}$ denotes the probability ratio for an action $a_t$ in state $s_t$, the clipped \ppo objective is:
\[
\mathcal{L}^{\text{\ppo}}(\theta) = \mathbb{E}_t \Big[
  \min\big( r_t(\theta) \, A_t,\, 
  \mathrm{clip}(r_t(\theta), 1-\epsilon, 1+\epsilon)\, A_t \big)
\Big],
\]
where $A_t$ is an estimator of the advantage function and $\epsilon$ is a hyperparameter controlling the allowable deviation from the previous policy. $A_t$ is computed using Generalized Advantage Estimation (\texttt{GAE})~\cite{schulman2018highdimensionalcontinuouscontrolusing} based on rewards and a learned value function. The clipping objective of \ppo restricts how drastically the updated policy distribution can diverge from the original policy. This moderation averts catastrophic shifts in language generation and preserves training stability.


\textbf{Policy Optimization with KL Penalty.}
During \rl fine-tuning with \ppo, the policy $\pi$ is optimized to maximize reward while staying close to the base model $\rho$. The modified reward function includes a KL divergence penalty:
\[
J(\pi) = \mathbb{E}_{(x,y) \sim \mathcal{D}} \left[ r(x,y) - \beta \, \mathrm{KL}\bigl(\pi(\cdot|x) \,\|\, \rho(\cdot|x)\bigr) \right],
\]
where $\beta$ controls the penalty strength. The KL term $\mathrm{KL}(\pi \,\|\, \rho)$ prevents over-optimization to the proxy reward $r(x,y)$ (i.e., reward hacking).


\insightbox{The KL penalty is a regularization, which ensure policy retains the base model’s linguistic coherence and avoids degenerate outputs.}
\subsubsection{Reinforcement Learning from Human Feedback (\rlhf) }
\label{sub_rlhf}
\rlhf~\cite{ouyang2022training} refines \llms through direct human preference signals, making them more aligned with human expectations. The process involves three main steps. First, \texttt{SFT} is performed on a pretrained model using high-quality labeled data to establish strong linguistic and factual capabilities. Second, a reward function $R$ is trained using human-annotated rankings of generated responses, allowing it to predict preferences and provide a scalar reward signal. Third, \ppo is employed in the \rlhf~\cite{ouyang2022training} pipeline by using human-provided preference scores (or rankings) to shape $R$ and thereby guide the policy updates. This ensures that the model prioritizes outputs aligned with human-preferred behavior. The robust performance under conditions of noisy or partial reward signals makes \ppo well-suited for text generation tasks, where large action spaces and nuanced reward definitions are common.
\subsubsection{Reinforcement Learning from AI Feedback (\texttt{RLAIF})}
\label{sub_rlaif}
\texttt{RLAIF}~\cite{lee2023rlaif} is an alternative to \rlhf that replaces human annotation with AI-generated feedback. Instead of relying on human-labeled preferences, \texttt{RLAIF} employs a secondary, highly capable language model to generate preference labels, which are then used to train a reward model. This reward model guides reinforcement learning-based fine-tuning of the target model.
\texttt{RLAIF} reduces the cost and time required for data collection by eliminating the need for human annotators. It enables large-scale model alignment without requiring extensive human intervention while maintaining high performance and alignment. Empirical studies indicate that \texttt{RLAIF} \cite{lee2023rlaif,Rafailov2024ScalingLF} is a scalable and efficient alternative to \rlhf, making it a promising direction for reinforcement learning-driven language model optimization.
\insightbox{The clipping mechanism constrains policy updates to remain within a safe trust region, which is crucial when dealing with complex, high-dimensional action spaces.}
\subsubsection{Trust Region Policy Optimization (\trpo)}
\label{sub_trpo}
\trpo~\cite{schulman2017trustregionpolicyoptimization} is another widely used policy optimization method, preceding \ppo and sharing its fundamental goal: improving stability in reinforcement learning updates. \trpo optimizes policy updates while ensuring they remain within a constrained trust region, measured by KL divergence.

Instead of using a clipped objective like \ppo, \trpo enforces a hard constraint on policy updates by solving the following optimization problem:
\[
\max_{\theta} \quad \mathbb{E}_{t} \left[\frac{\pi_{\theta}(a_t \mid s_t)}{\pi_{\theta_{\text{old}}}(a_t \mid s_t)} A_t \right]
\]
subject to the constraint:
\[
\mathbb{E}_{t} \left[ D_{KL} \left(\pi_{\theta_{\text{old}}}(\cdot \mid s_t) \| \pi_{\theta}(\cdot \mid s_t) \right) \right] \leq \delta.
\]
where $\delta$ is a hyperparameter that controls how much the new policy can diverge from the old one.

Unlike \ppo, which approximates this constraint using clipping, \trpo directly solves a constrained optimization problem, ensuring each update does not move too far in policy space. However, solving this constrained problem requires computationally expensive second-order optimization techniques like conjugate gradient methods, making \trpo less efficient for large-scale models like \llms. In practice, \ppo is preferred over \trpo due to its simplicity, ease of implementation, and comparable performance in large-scale applications like \rlhf. However, \trpo remains an important theoretical foundation for stable policy optimization in deep reinforcement learning. 

\subsubsection{Direct Preference Optimization (\dpo)}
\label{sec_dpo}
\dpo~\cite{guo2024direct} is a recently proposed method for training \llms from human preference data without resorting to the traditional \rl loop (as in \rlhf with \ppo). Instead of learning a separate reward function and then running policy-gradient updates, \dpo directly integrates human preference signals into the model's training objective. So instead of the above \ppo objective, \dpo instead constructs an objective that directly pushes up the probability of a chosen (preferred) response ($y^+$) while pushing down the probability of a less-preferred response  ($y^-$), all within a single log-likelihood framework. Rather than bounding policy changes with clip, the \dpo loss uses the difference between log probabilities of `winning'  vs. `losing' responses. This explicitly encodes the user’s preference in the updated parameters.

\input{figures/PPO_comp_fig}

Here, $\pi_\theta$ is the learnable policy, $\pi_{\text{ref}}$ is a reference policy (often the SFT-trained model), $\sigma(\cdot)$ is the sigmoid function, $\beta$ is a scaling parameter, and $\mathcal{D}_{\text{train}}$ is a dataset of triplets $(x, y^+, y^-)$ where $y^+$ is the preferred output over $y^-$.
\[
\mathcal{L}^{\mathrm{DPO}}(\theta) 
= \mathbb{E}_{((x,y^+), y^-) \sim \mathcal{D}_{\mathrm{train}}} \Bigl[
  \sigma\!\Bigl(
    \beta \log \frac{\pi_\theta(y^+ \mid x)}{\pi_{\mathrm{ref}}(y^+ \mid x)} \]
\[
    - \beta \log \frac{\pi_\theta(y^- \mid x)}{\pi_{\mathrm{ref}}(y^- \mid x)}
  \Bigr)
\Bigr].
\]
The key insight is that an \llm can be treated as a “hidden reward model”: we can reparameterize preference data so that the model’s own log probabilities reflect how preferable one response is over another. By directly adjusting the log-likelihood of more-preferred responses relative to less-preferred ones, \dpo sidesteps many complexities of \rl-based methods (e.g., advantage functions or explicit clipping). 
\insightbox{The advantage function $A_\phi = V_\phi(\mathbf{s}_{t+1}) - V_\phi(\mathbf{s}_t)$ quantifies per-step contributions, critical for identifying key reasoning errors. This granularity is lost in \dpo, which treats entire trajectories uniformly.}  
\noindent\textbf{Perplexity Filtering for Out-of-Distribution Data.}
To ensure \dpo training data is on-distribution (aligned with $\rho$), responses are filtered using perplexity. The perplexity of a response $y = (y_1, y_2, \dots, y_T)$ is defined as:
\[
\text{PP}(y) = \exp\left(-\frac{1}{T}\sum_{i=1}^{T}\log P_\rho(y_i \mid y_{<i})\right),
\]
where $y_i$ is the $i$-th token. Only responses with perplexity below a threshold (e.g., the 95th percentile of $\rho$-generated responses) are retained.
\insightbox{The advantage function remains a core concept to determine which actions (token choices) are better than the baseline at each step.
}

\subsubsection{Offline Reasoning Optimization (\texttt{OREO})}  
\label{sec_oreo}
\texttt{OREO}~\cite{oreo} is an offline reinforcement learning method designed to enhance \llms' multi-step reasoning by optimizing the soft Bellman equation~\cite{bellman1957markovian}. Unlike \dpo, which relies on paired preference data, \texttt{OREO} uses sparse rewards based on final outcomes (e.g., correctness of reasoning chains) and jointly trains a policy model $\pi_\theta$ and a value function $V_\phi$ for fine-grained credit assignment.  
The core objective minimizes the inconsistency in the soft Bellman equation:  
\[
V_\phi(\mathbf{s}_t) - V_\phi(\mathbf{s}_{t+1}) = r(\mathbf{s}_t, \mathbf{a}_t) - \beta \log \frac{\pi_\theta(\mathbf{a}_t \mid \mathbf{s}_t)}{\pi_{\text{ref}}(\mathbf{a}_t \mid \mathbf{s}_t)},
\]  
where $\mathbf{s}_{t+1} = f(\mathbf{s}_t, \mathbf{a}_t)$ is the next state, $r$ is the sparse reward, and $\beta$ controls KL regularization. The policy and value losses are:  
\[
\mathcal{L}_V(\phi) = \frac{1}{T} \sum_{t=0}^{T-1} \left( V_\phi(\mathbf{s}_t) - R_t + \beta \sum_{i \geq t} \log \frac{\pi_\theta(\mathbf{a}_i \mid \mathbf{s}_i)}{\pi_{\text{ref}}(\mathbf{a}_i \mid \mathbf{s}_i)} \right)^2,
\]  
\[
\mathcal{L}_\pi(\theta) = \frac{1}{T} \sum_{t=0}^{T-1} \left( V_\phi(\mathbf{s}_t) - R_t + \beta \log \frac{\pi_\theta(\mathbf{a}_t \mid \mathbf{s}_t)}{\pi_{\text{ref}}(\mathbf{a}_t \mid \mathbf{s}_t)} \right)^2 + \alpha \mathcal{L}_{\text{reg}},
\]  
where $\mathcal{L}_{\text{reg}}$ penalizes deviations from $\pi_{\text{ref}}$, and $\alpha$ balances regularization.  

\insightbox{\texttt{OREO’s} explicit value function enables test-time beam search (e.g., selecting high-value reasoning steps) and iterative training, where failed trajectories refine the policy. This contrasts with \dpo implicit value function, which lacks stepwise credit assignment.}  

\warningbox{\texttt{OREO’s} computational cost scales with trajectory length and value-model training. While effective for math/agent tasks, its generalization to broader domains (e.g., coding) requires validation. Iterative training also demands careful data curation to avoid overfitting to failure modes.}
\subsubsection{Group Relative Policy Optimization (\grpo)}
\label{sec_grpo}
\grpo~\cite{shao2024deepseekmath} simplifies the \ppo framework by eliminating the need for a separate value function. Instead, \grpo estimates the baseline from the average reward of multiple sampled outputs for the same question.
The primary contribution in \grpo is that it removes the need for a separate value model (critic model) and instead estimates the baseline reward from a group of sampled \llm outputs. 
This significantly reduces memory usage and stabilizes policy learning. 
The approach also aligns well with how reward models are trained, i.e., by comparing different \llm-generated outputs rather than predicting an absolute value.

For each question \(q\), \grpo samples a group of outputs \(\{ o_1, o_2, \dots, o_G \}\) from the old policy \(\pi^{old}_\theta\). A reward model is used to score each output in the group, yielding rewards \(\{ r_1, r_2, \dots, r_G \}\). The rewards are normalized by subtracting the group average and dividing by the standard deviation:
\[
\bar{r}_i = \frac{r_i - \mathrm{mean}(r)}{\mathrm{std}(r)}.
\]
The advantage \(\hat{A}_{i,t}\) for each token in the output is set as the normalized reward \(\bar{r}_i\).


\grpo first samples a question \(q \sim P(Q)\) and then samples \(G\) outputs \(\{o_i\}_{i=1}^G\) from \(\pi_\theta^{old}(O \mid q)\). Define the per-output objective as
\begin{align*}
J(o_i, \theta, q) = \frac{1}{|o_i|}\sum_{t=1}^{|o_i|} \Biggl( & \min\Big\{ r_{\text{ratio}, i, t}\, \hat{A}_{i,t},\, \\
& \quad \text{clip}\bigl(r_{\text{ratio}, i,t}, 1-\epsilon, 1+\epsilon\bigr)\, \hat{A}_{i,t} \Big\} \nonumber \\
& \quad - \beta\, D_{KL}\Big[\pi_\theta \,\|\, \pi_{ref}\Big] \Biggr).
\end{align*}
Then, the GRPO objective becomes
\[
J_{GRPO}(\theta) = \mathbb{E}_{q \sim P(Q)} \left[ \frac{1}{G} \sum_{i=1}^G J(o_i, \theta, q) \right],
\]
where the probability ratio is defined as
\[
r_{\text{ratio}, i, t} \triangleq \frac{\pi_\theta(o_{i,t} \mid q, o_{i,<t})}{\pi_\theta^{old}(o_{i,t} \mid q, o_{i,<t})}.
\]
where $\epsilon$ is a clipping hyperparameter akin to \ppo, and $\beta$ adjusts the KL-divergence penalty encouraging the new policy $\pi_\theta$ not to deviate excessively from a reference policy $\pi_{\mathrm{ref}}$, which is typically the initial supervised fine-tuned (\texttt{SFT}) model~\cite{mukobi2023superhf, li2023task}. \grpo can be applied in two modes: outcome supervision and process supervision.

\textbf{Outcome Supervision:} Provides a reward only at the end of each output. The advantage 
\(\hat{A}_{i,t}\) for all tokens in the output is set as the normalized reward \(\bar{r}_i\).
\[
\bar{r}_i = \frac{r_i - \mathrm{mean}(r)}{\mathrm{std}(r)}.
\]

\textbf{Process Supervision:} Provides a reward at the end of each reasoning step. The advantage 
\(\hat{A}_{i,t}\) for each token is calculated as the sum of the normalized rewards from the following steps:
\[
\hat{A}_{i,t} = \sum_{\text{index}(j) \geq t} \bar{r}_{i,\text{index}(j)},
\]
where \(\text{index}(j)\) is the end token index of the \(j\)-th step.\\
Overall, \grpo serves as an efficient alternative to classic actor-critic frameworks in \textbf{DeepSeekR1}~\cite{guo2025deepseek} by leveraging group-level advantages, thereby reducing training costs without sacrificing the capacity to distinguish fine-grained differences among candidate responses.
\insightbox{Fine-grained per-step rewards enable the model to effectively identify and reinforce high-quality responses, boosting overall performance in complex, multi-step reasoning tasks.}
\subsubsection{Multi-Sample Comparison Optimization}
Instead of relying solely on single-pair comparisons, multi-sample comparison optimization~\cite{wang2024preferenceoptimizationmultisamplecomparisons} approach compares multiple responses simultaneously to promote diversity and mitigate bias. Specifically, given a set of responses $\{ y_1, y_2, \dots, y_n \}$ for a query $x$, the probability of observing the ranking $y_1 > y_2 > \dots > y_n$ is determined by the product
\[
    P(y_1 > y_2 > \dots > y_n)
    \;=\;
    \prod_{i}
    \frac{e^{R(x, y_i)}}
    {\sum_{j} e^{R(x, y_j)}}.
\]
In this formulation, each response $y_i$ is jointly evaluated in the context of all other responses, ensuring that comparisons are not isolated pairwise events but rather part of a broader ranking framework that helps capture more nuanced preferences and reduces potential biases.

\subsection{Pure \rl Based LLM Refinement}

The work from Guo et al. (2025) \cite{guo2025deepseek} introduces two main models: \emph{DeepSeek-R1-Zero} and \emph{DeepSeek-R1}. 
\begin{itemize}
    \item \textbf{DeepSeek-R1-Zero} operates with a purely Reinforcement Learning approach, excluding any \texttt{SFT}.
    \item \textbf{DeepSeek-R1} incorporates \emph{cold-start} data and applies a multi-stage training pipeline.
\end{itemize}


The methodology encompasses several steps (See Figure \ref{fig:RL_overview} in GRPO for main steps): collecting cold-start data, performing \rl training, carrying out \texttt{SFT}, using distillation to transfer knowledge to smaller models, and addressing specific challenges such as language mixing and readability.
This multi-stage pipeline ensures robustness and alignment with human preferences, while distillation enables efficient deployment of smaller models without significant performance loss.

\subsubsection{Cold-Start \rl Phase}
The process begins with a \emph{cold-start} \rl phase, where a small amount of curated data is gathered to fine-tune an initial, or \emph{base}, model. Following this preliminary fine-tuning, \rl is conducted—often via algorithms like \grpo until convergence.
 The cold-start phase is critical for stabilizing the model before full \rl training, preventing instability that can arise from purely \rl-driven updates.
 The \emph{cold-start data preparation} focuses on capturing human-readable reasoning patterns to prevent instability from purely \rl-driven updates. This step generates CoT-style examples with consistent $<reasoning\_process>$ and $<summary>$ fields, usually involving thousands of carefully curated samples.
Structured \texttt{CoT} formats and consistent fields ensure clarity and robustness in the model's reasoning outputs, reducing errors and improving interpretability~\cite{wei2022chain, suzgun2022challenging, cpo, zhang2022automatic}.
\insightbox{Providing \texttt{CoT} reasoning traces before \rl training establishes a stronger foundation for reasoning tasks, enhancing both robustness and interpretability of outputs.}
\subsubsection{Rejection Sampling and Fine-tuning}
This concept is also used in WebGPT~\cite{nakano2021webgpt}. Once \rl stabilizes, a \emph{rejection sampling} mechanism is employed to generate high-quality responses that are subsequently filtered for correctness, clarity, and other quality metrics. These filtered responses are then blended with additional datasets to produce a new, larger corpus for Supervised Fine-Tuning.
Rejection sampling ensures that only high-quality outputs are used for further training, enhancing the model's overall performance and reliability.
After \rl converges for high-stakes reasoning tasks, \emph{rejection sampling} is used to filter a large number of generated outputs, expanding the training set. These newly generated reasoning examples (potentially up to hundreds of thousands in quantity) are mixed with existing \texttt{SFT} data to create a combined dataset of substantial size (often around 800k samples).
Rejection sampling and dataset expansion significantly enhance the model's coverage of general tasks while preserving its reasoning proficiency.

\subsubsection{Reasoning-Oriented RL}
The \emph{reasoning-oriented \rl} leverages \grpo~\cite{shao2024deepseekmath}, which samples a group of outputs from the current policy and computes rewards and advantages for each output. Rewards may be computed via rule-based checks, e.g., ensuring correct solutions in math or code tasks, enforcing structured \texttt{CoT} tags, and penalizing undesired language mixing.
\grpo group-based sampling and reward computation ensure that the model prioritizes high-quality, structured outputs, enhancing its reasoning capabilities.
\begin{figure}[t]
\centering
    \includegraphics[width=\columnwidth]{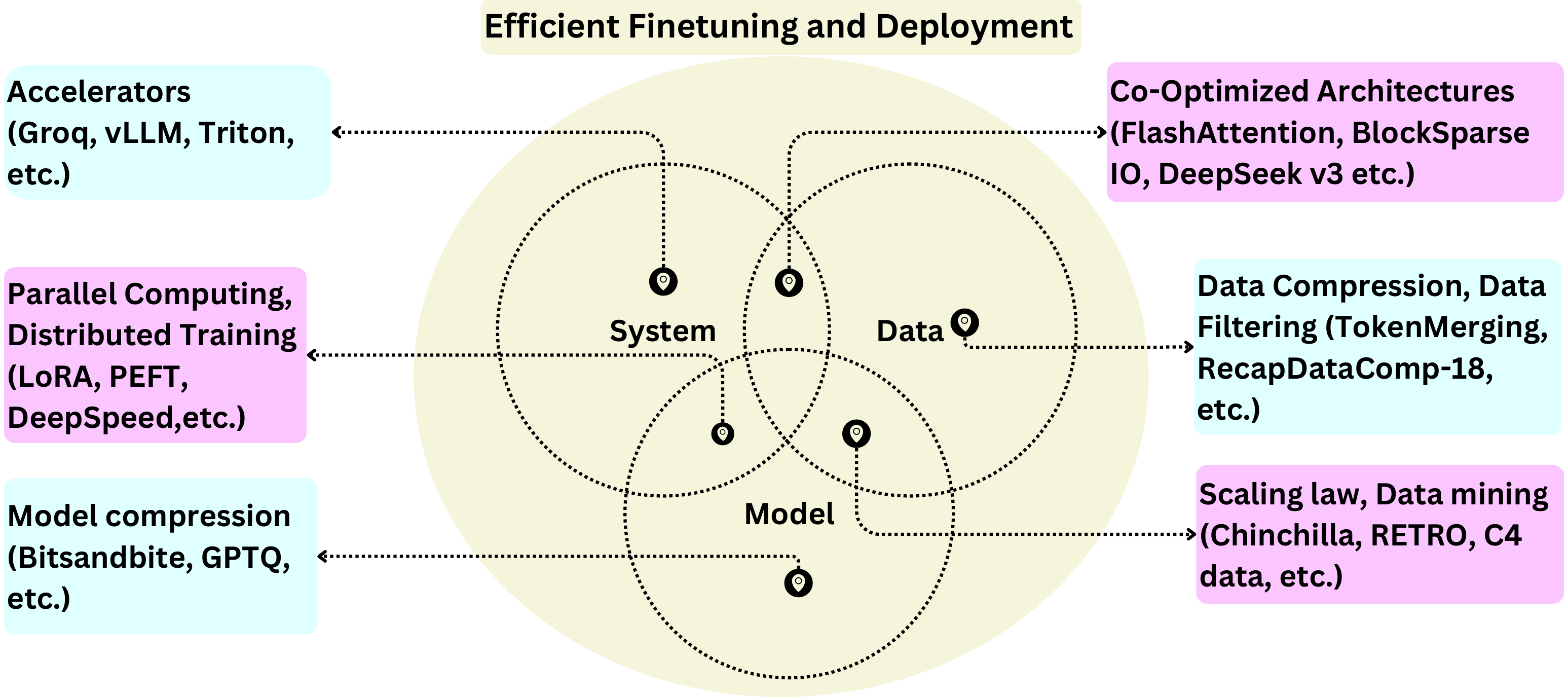}
    \caption{\small This Venn diagram illustrates the interplay between System, Data, and Model for efficient finetuning and deployment. It covers strategies like accelerators (Groq, vLLM), adaptation (\lora, \texttt{PEFT}), co-optimized architectures (FlashAttention), data compression (TokenMerging), scaling laws (Chinchilla), and model compression (\texttt{GPTQ}) to boost performance and scalability.}
    \label{fig:sft}
\end{figure}

\subsubsection{Second \rl Stage for Human Alignment}
A \emph{second \rl stage} further aligns the model with broader human preferences (helpfulness, harmlessness, creativity, etc.) by introducing additional reward signals and prompt distributions. The second \rl stage ensures the model aligns with human values, making it more versatile and contextually aware.
After re-training the base model on this combined dataset, a second round of \rl can be conducted to align the model more closely with human preferences (e.g., for helpfulness and harmlessness).
This \rl stage fine-tunes the model to better align with human values, ensuring outputs are not only accurate but also contextually appropriate.

\subsubsection{Distillation for Smaller Models}
Finally, \emph{distillation} techniques are used to transfer the refined capabilities of the main model to smaller architectures, enabling more efficient deployments without sacrificing much performance.
It allows smaller models to inherit advanced reasoning capabilities, making them competitive on challenging benchmarks without the computational costs of full-scale \rl training.
Finally, \emph{distillation} plays a pivotal role: the top-performing model, \emph{DeepSeek-R1}~\cite{guo2025deepseek}, serves as a teacher to smaller architectures (e.g., Qwen or Llama families, ranging from 1.5B to 70B parameters). This transfer allows the smaller models to inherit advanced reasoning capabilities, making them competitive on challenging benchmarks without incurring the computational costs of full-scale \rl training.

\insightbox{Distillation democratizes advanced reasoning capabilities, enabling smaller models to achieve competitive performance with reduced computational overhead.}

\section{Supervised Finetuning in \llms}
\label{sec_fintune}
As shown in Figure~\ref{fig:RL_overview}, finetuning forms a basic component of \llm post-training recipes. In this section, we summarize the different types of \llm fine-tuning mechanisms. 

\subsection{Instruction finetuning}
In instruction finetuning, a model is trained on curated pairs of instruction (prompt) and response (completion). The main goal is to guide the \llm to follow a user‐provided instruction accurately and helpfully, regardless of the task domain. This usually involves compiling large, diverse instruction‐response datasets covering many task types (e.g., summarization, QA, classification, creative writing). Models such as T0 \cite{sanh2021multitask}, FLAN \cite{wei2021finetuned}, Alpaca \cite{alpaca}, Vicuna \cite{vicuna2023} and Dolly \cite{DatabricksBlog2023DollyV2} demonstrate how instruction‐finetuned \llms can outperform base models on zero‐shot or few‐shot tasks by virtue of their enhanced instruction‐following abilities.

\subsection{Dialogue (Multi-turn) Finetuning}
Some \llms undergo dialogue‐style finetuning to better handle multi‐turn conversations. Different from instruction tuning described above, here the data takes the form of a continuous dialogue (multi-turn conversations) instead of a single prompt-response pair. In this approach, training data consists of chat transcripts with muliple user queries and system responses, ensuring the model learns to maintain context across turns and produce coherent replies. Models like LaMDA \cite{thoppilan2022lamda} and ChatGPT \cite{openai2023gpt} highlight how dialogue‐tuned \llms can feel more interactive and context‐aware. While dialogue finetuning can overlap with instruction finetuning (because many instructions come in a chat format), specialized conversation data often yields more natural, multi‐turn user experiences.

\subsection{CoT Reasoning finetuning}
Chain‐of‐Thought (\texttt{CoT}) reasoning finetuning teaches models to produce step‐by‐step reasoning traces instead of just final answers. 
By exposing intermediate rationales or \emph{thoughts}, \texttt{CoT} finetuning can improve both interpretability and accuracy on complex tasks (e.g., math word problems, multi‐hop QA). In practice, \texttt{CoT} finetuning uses supervised reasoning annotations (often handcrafted by experts) to show how a solution unfolds. Notable early work includes Chain‐of‐Thought Prompting \cite{wei2022chain} and Self‐Consistency \cite{wang2023selfconsistency}, which initially applied the idea to prompting; subsequent efforts (e.g., Chain‐of‐Thought Distillation \cite{magister2022teaching}) adapt it to a full finetuning or student‐teacher paradigm.
These efforts have also been extended to the multimodal domain, e.g., LlaVA-CoT \cite{xu2024llavacot} and LlamaV-o1 \cite{thawakar2025llamavo1} where image, QA and CoT reasoning steps are used in \llm finetuning.

\input{tables/methods.tex}

\subsection{Domain‐Specific (Specialized) Finetuning}
When an \llm needs to excel in a specific domain (e.g., biomedicine, finance, or legal), domain‐specific finetuning is used. Here, a curated corpus of domain‐relevant text and labeled examples is employed to finetune the \llm. For instance, BioGPT \cite{luo2022biogpt} and BiMediX \cite{pieri2024bimedix} specialize in biomedical literature, FinBERT \cite{yang2020finbert} for financial texts, ClimatGPT \cite{thulke2024climategpt,mullappilly2023arabic} for climate and sustainability and CodeT5 \cite{wang2021codet5} for code understanding. Supervised finetuning in these domains often includes classification, retrieval, or QA tasks with domain‐specific data, ensuring the model’s parameters adapt to the specialized language and concepts of the field.
Domain-specific finetuning is also extended to vision-language models such as, \cite{kuckreja2024geochat} finetuned on remote sensing imagery, \cite{mullappilly2024bimedix2} on medical imaging modalities, \cite{maaz2023video,lin2023video,zhang2023video} on spatiotemporal video inputs, and \cite{han2023chartllama} adapted for chart understanding. 

\subsection{Distillation‐Based Finetuning}
Large `teacher' models are sometimes used to produce labeled data or rationales, which a smaller `student' model finetunes on, this is generally called knowledge distillation \cite{zhu2024survey,wan2023efficient}. In the context of \llms, \texttt{CoT} Distillation \cite{magister2022teaching} is one example where a powerful teacher \llm generates intermediate reasoning steps, and the student \llm is finetuned to reproduce both the final answer and the reasoning chain. Step-by-step distillation \cite{hsieh2023distilling} generates descriptive rationales alongside final answers to train smaller models through distillation with smaller datasets.
This approach can yield lighter, faster models that retain much of the teacher’s performance, even in zero‐shot or few‐shot tasks \cite{gu2023minillm}.

\subsection{Preference and Alignment SFT}
While RLHF is not purely supervised, it starts with a supervised \emph{preference} or \emph{alignment} finetuning stage. This stage uses human‐labeled or human‐ranked examples to teach the model about desirable vs. undesirable outputs (e.g., safe vs. toxic). By training on these explicit preferences, the model becomes more aligned with user values, reducing harmful or off‐topic completions. Works like InstructGPT \cite{ouyang2022training} illustrate how supervised preference data is critical before reward model training and \rl updates begin.

\subsection{Efficient Finetuning}
\label{sec:efficient_finetuning}

Fully finetuning a \llm can be computationally and memory‐intensive, particularly as model sizes grow into the tens or hundreds of billions of parameters. To address these challenges, parameter‐efficient finetuning (\texttt{PEFT}) techniques introduce a small set of trainable parameters or learnable prompts while leaving most of the model weights frozen. Approaches such as LoRA~\cite{hu2021lora}, Prefix Tuning \cite{li2021prefix}, and Adapters \cite{houlsby2019parameterefficienttransferlearningnlp}  exemplify this strategy by injecting lightweight modules (or prompts) in specific layers, thus significantly reducing the memory footprint.

Figure~\ref{fig:sft} illustrates how these techniques fit into a broader ecosystem that involves system‐level optimizations, data management, and evaluation strategies for \llms. In particular, \texttt{PEFT} approaches can be combined with quantization and pruning methods~\cite{frantar-sparsegpt, dettmers2023qlora} to further minimize memory usage and compute overhead, enabling finetuning on smaller GPUs or even consumer‐grade hardware. For instance, \texttt{QLoRA} unifies 4‐bit quantization with low‐rank adaptation, while \texttt{BitsAndBytes} provides 8‐bit optimizers to make \llm training more practical in constrained environments (Table~\ref{tab:llm_tools}).

Moreover, these \texttt{PEFT} methods still require supervised data to guide the adaptation process, but the reduction in the number of trainable parameters makes it more feasible to use in‐domain or task‐specific datasets. This is especially valuable for specialized domains (e.g., medical or software development), where data might be limited or expensive to annotate. As shown in Table~\ref{tab:llm_tools}, \texttt{PEFT (HF)} integrates several of these approaches (LoRA, prefix tuning, and more) into a single library, streamlining deployment in both research and production settings.

\insightbox{Combining efficient tuning designs like \lora and \texttt{QLoRA} with system and data optimizations (Figure~\ref{fig:sft}) enables cost-effective LLM adaptation for tasks like domain-specific text generation,
without expensive full fine-tuning.}

\section{Test-time Scaling Methods}
\label{sec_tts}
\begin{figure}
    \centering
    \includegraphics[width=\columnwidth]{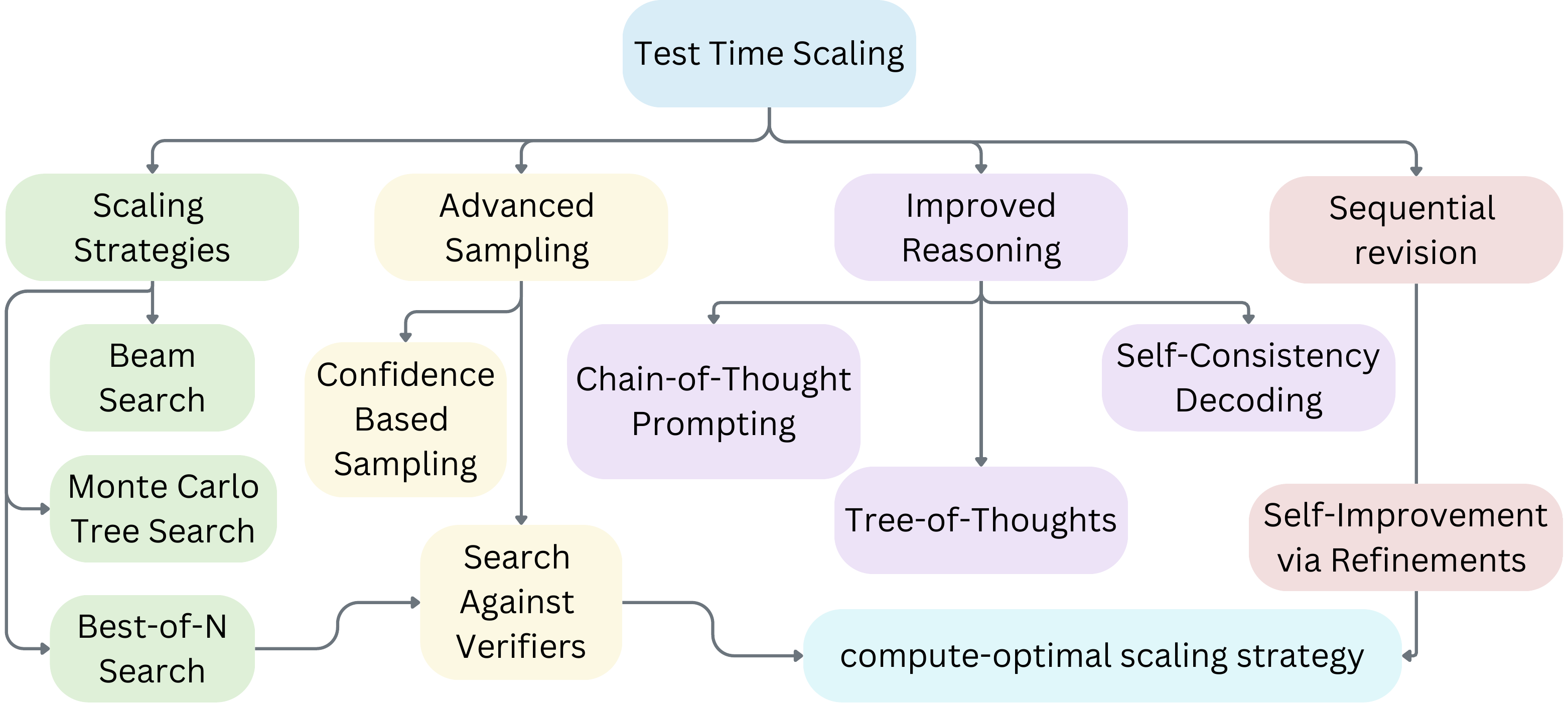}
    \caption{An overview of Test-time Scaling methods: parallel scaling, sequential scaling, and search-based methods. It also shows how they integrate into a compute-optimal strategy.}
    \label{fig:tts}
\end{figure}

While \rl fine-tunes the model's policy, test-time scaling (\texttt{TTS}) enhances reasoning during inference typically without model updates. Figure \ref{fig:tts} presents a taxonomy of \texttt{TTS} methods, categorizing them based on their underlying techniques.
\subsection{Beam Search} 
Beam search was first introduced in the context of speech recognition \cite{lowerre1976harpy}. 
It gained prominence as a decoding strategy for sequence models and was later adopted in neural machine translation and speech systems \cite{graves2012sequence}.
With the popularity of \llms, this algorithm has been used for approximate search in many text generation tasks.

The concept of Beam search is similar to pruned \emph{breadth-first search}, where top N highest-probability partial sequences (the `beam') are kept at each step, discarding lower-probability paths. 
By limiting the beam width (N), it manages the exponential search space while aiming to find a near-optimal sequence.
These beams are expanded at each decoding step to find multiple probable paths. 
In reasoning \llms, such paths allow us to systematically explore multiple reasoning chains in parallel, focusing on the most promising ones. 
This ensures that high-likelihood reasoning steps are considered, which can improve the chances of finding a correct and coherent solution compared to greedy decoding. 
It has traditionally been used in tasks such as translation, summarization, and code generation, where the goal is a highly probable correct sequence \cite{wu2025survey}.

While modern \llms often favor stochastic sampling (e.g., temperature sampling) to promote diversity in generated text, beam search is still a valuable technique for structured reasoning problems. 
For example, the Tree-of-Thoughts framework \cite{yao2024tree} allows plugging in different search algorithms to explore a tree of possible `\emph{thoughts}' or reasoning steps; usually a beam search (with beam width $b$) is used to maintain the $b$ most promising states at each reasoning step.
Here, beam search is used to systematically explore solution steps for tasks like mathematical puzzles and planning problems, pruning less promising reasoning branches and thus improving the model’s problem-solving accuracy.  
Beam search remains a strong baseline for test-time reasoning when one wants the model to output the single most likely reasoning path or answer under the model’s learned distribution. 

\subsection{Best-of-N Search (Rejection Sampling)}
Best-of-N (\texttt{BoN})~\cite{sun2024fastbestofndecodingspeculative} search generates N candidate outputs (usually via sampling) and then picks the best one according to a chosen criterion (e.g., a reward model or the model’s own likelihood) \cite{askell2021general,glaese2022improving,stiennon2020learning}. 
Conceptually, this is an application of rejection sampling: one draws multiple samples and rejects all but the top-rated result.
Unlike Beam Search \cite{lowerre1976harpy,graves2012sequence}, which incrementally expands and prunes partial hypotheses, BoN simply samples full solutions independently, allowing for greater diversity but at a higher computational cost. 
Beam Search systematically aims for the most probable sequence, while BoN may capture high-quality but lower-probability solutions through brute-force sampling.
\insightbox{Beam search (effective for harder questions) outperforms best-of-N sampling at low compute budgets, while best-of-N scales better for easier tasks.}
During \llm inference, \texttt{BoN} is used to enhance correctness or alignment without retraining the model. 
By sampling multiple answers and selecting the top candidate (e.g., via a reward model or a checker), \texttt{BoN} effectively boosts accuracy on tasks like QA or code generation. 
\texttt{BoN} is easy to understand and implement and is almost hyper-parameter-free, with N being the only parameter that can be adjusted at inference. 
In reinforcement learning contexts, \texttt{BoN} sampling can serve as a baseline exploration mechanism i.e., to generate many rollouts, pick the best outcome according to the learned reward, and proceed, although at increased computational overhead. 
\textit{OpenAI’s} WebGPT used \texttt{BoN} to pick the best response via a reward model, yielding strong QA performance \cite{nakano2021webgpt}. 
BoN is also used as a simple alignment method that is highly competitive with other post-training techniques e.g., \rlhf~\cite{ouyang2022training} and \dpo \cite{dubois2024alpacafarm}.
Studies have shown \texttt{BoN} can approach or match \rlhf results when guided by a sufficiently robust reward model \cite{gao2023scaling,yang2024asymptotics}. Alternatives such as speculative rejection \cite{sun2024fast} build on this idea and utilize a better reward model to improve efficiency.
The studies also highlight issues of \emph{reward hacking} if the (proxy) reward function used for \texttt{BoN} is imperfect \cite{hilton2022measuring} or instability issues if the N parameter gets very large.
\insightbox{Choice of either process reward models with beam search vs best-of-N depends on the difficulty and compute budget.}
\subsection{Compute-Optimal Scaling}
The Compute-Optimal Scaling Strategy (\texttt{COS}) \cite{snell2024scaling} is a dynamic method designed to allocate computational resources efficiently during inference in \llms, optimizing accuracy without unnecessary expense. Instead of applying a uniform sampling strategy across all inputs, this approach categorizes prompts into five difficulty levels—ranging from easy to hard—either by leveraging oracle difficulty (ground-truth success rates) or model-predicted difficulty (e.g., verifier scores from Preference Ranking Models). Once categorized, the strategy adapts compute allocation: easier prompts undergo sequential refinement, where the model iteratively refines its output to improve correctness, while harder prompts trigger parallel sampling or beam search, which explores multiple response variations to increase the likelihood of finding a correct solution. This dual approach balances exploration (for challenging inputs) and refinement (for near-correct responses), ensuring optimal performance per unit of computational effort. Remarkably, this method achieves four times lower compute usage than traditional best-of-N sampling while maintaining equivalent performance. The key insight is that by matching computational strategy to problem difficulty, it avoids wasted resources on trivial cases while ensuring sufficient sampling diversity for complex tasks. In essence, it functions as a “smart thermostat” for \llm inference, dynamically adjusting computational effort in response to input complexity, leading to a more efficient and cost-effective deployment of large-scale language models.
\insightbox{COS achieves 4× efficiency gains over best-of-N baselines by optimally balancing sequential/parallel compute.
Beam search + revisions outperform larger models on easy/intermediate questions.}
\subsection{Chain-of-thought prompting}\label{cotprompt}
\texttt{CoT} prompting induces \llms to produce intermediate reasoning steps rather than jumping directly to the final answer. 
By breaking down problems into logical sub-steps, \texttt{CoT} taps into a model’s latent ability to perform multi-step inferences, significantly improving performance on tasks like math word problems, logical puzzles, and multi-hop QA.

Wei et al.~\cite{wei2022chain} demonstrated CoT’s effectiveness on arithmetic and logic tasks, showing large gains over direct prompting. Kojima et al.~\cite{wang2023plan} introduced Zero-Shot \texttt{CoT}, revealing that even adding a simple phrase like “Let’s think step by step” can trigger coherent reasoning in sufficiently large models. 
Subsequent works (e.g., Wang et al., 2022 \cite{wang2023selfconsistency}) combined \texttt{CoT} with sampling-based strategies (Self-Consistency) for even higher accuracy.
As described in Sec. \ref{cotprompt}, \texttt{CoT} format data have also been used for \texttt{SFT} and are shown to help reshape the model responses to be more step-by-step.
\insightbox{Fine-tuning models to revise answers sequentially allows them to build on previous attempts, improving accuracy over time. This approach is particularly effective for easier questions, while parallel sampling (exploration) proves more beneficial for harder ones.}
\subsection{Self-Consistency Decoding}
Self-Consistency is a decoding strategy introduced by Wang et al. \cite{wang2022self}. 
It was proposed as an alternative to simple greedy decoding for chain-of-thought prompts.
It built upon the idea of sampling multiple distinct reasoning paths for a question and was the first to show that marginalizing over those paths can significantly improve accuracy on arithmetic and reasoning problems. 
In other words, it allows the model to think in many ways and then trust the consensus, which improves correctness in many reasoning scenarios.

The self-consistency method works by sampling a diverse set of reasoning chains from the model (via prompt engineering to encourage different \texttt{CoTs}, and using temperature sampling) and then letting the model output a final answer for each chain. 
Instead of trusting a single chain, the method selects the answer that is most consistent across these multiple reasoning paths, effectively a \emph{majority vote} or \emph{highest probability answer} after marginalizing out the latent reasoning. 
The intuition is that if a complex problem has a unique correct answer, different valid reasoning paths should converge to that same answer. 
By pooling the outcomes of many chains, the model can “decide” which answer is most supported. 
In application, one might sample, e.g., 20 \texttt{CoTs} for a math problem and see what final answer appears most frequently; that answer is then taken as the model’s output. 
This approach turns the one-shot \texttt{CoT} process into an ensemble where the model cross-verifies its answers. 
It is especially useful for arithmetic and commonsense reasoning tasks where reasoning diversity helps.
\insightbox{Smaller models with test-time compute can outperform much larger models in certain scenarios.}
Self-consistency is often combined with other methods: e.g., sampling multiple chains and then applying a verifier to the most common answer. Its strength lies in requiring no new training, only extra sampling, making it a popular test-time scaling strategy to obtain more reliable answers from \llms.
It has also inspired other variants, e.g., Universal Self-Consistency \cite{chen2024universal} extend the original  idea (which worked only with majority vote on single final answer) to more general generation tasks such as summarization and open-ended QA. 

\subsection{Tree-of-thoughts}
\texttt{ToT} framework \cite{yao2024tree}  generalizes the chain-of-thought approach by allowing the model to branch out into multiple possible thought sequences instead of following a single linear chain. It thus formulates the problem of language-model reasoning as a tree search, drawing on classic AI search methods inspired by human problem-solving \cite{newell1966analysis,newell1972human}.
\begin{figure*}
    \centering
    \includegraphics[width=\linewidth]{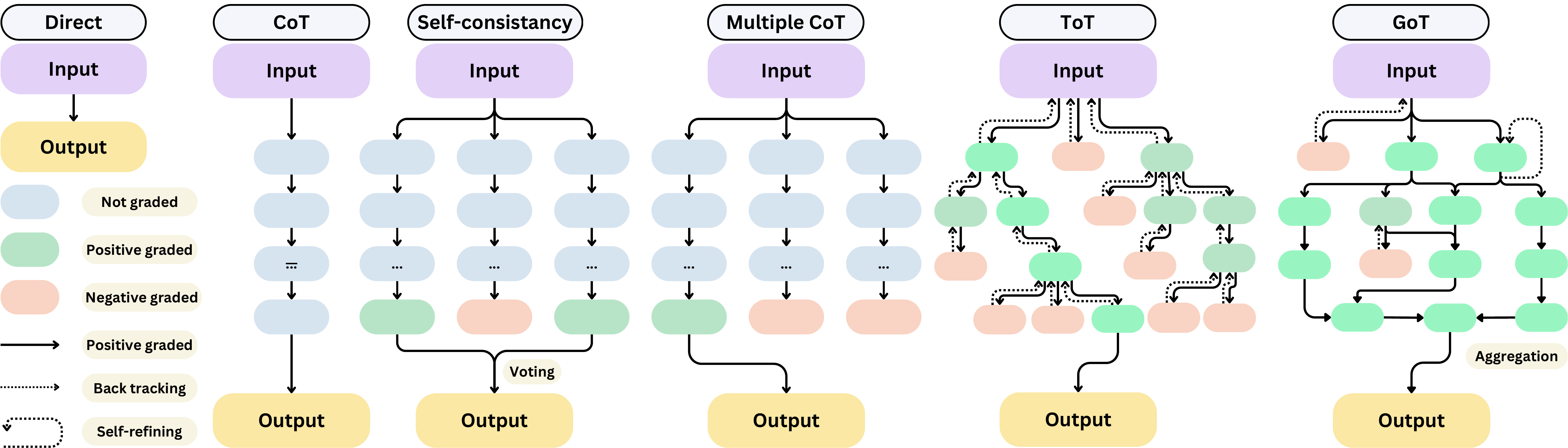}
    \caption{This figure compares reasoning strategies in \llms, evolving from Direct Prompting, which maps input to output without reasoning, to more structured approaches. Chain-of-Thought (\texttt{CoT}) introduces step-by-step reasoning, while Self-Consistency (\texttt{CoT-SC}) generates multiple \texttt{CoT} paths and selects the most frequent answer. Multiple \texttt{CoTs} explores diverse reasoning paths independently. Tree-of-Thoughts (\texttt{ToT}) structures reasoning as a tree, enabling backtracking and refinement, whereas Graph-of-Thoughts (\texttt{GoT}) generalizes this by dynamically aggregating and connecting thoughts. The legend deciphers key mechanisms like grading, backtracking, and self-refinement, crucial for optimizing reasoning efficiency.}
    \label{fig:tts_thoughts}
\end{figure*}
Tree of Thoughts treats intermediate reasoning steps as “nodes” in a search tree and uses the language model to expand possible next steps (thoughts) from a given state.
Rather than sampling one long reasoning path, the model explores a tree of branching thoughts and can perform lookahead and backtracking. 
At each step, the \llm might generate several candidate next thoughts, and a heuristic or value function evaluates each partial solution state. 
Then a search algorithm (e.g., depth-first, breadth-first, beam search) navigates this tree, deciding which branches to explore further.
This approach allows systematic exploration of different reasoning strategies: if one path leads to a dead-end, the model can return to an earlier state and try a different branch (unlike standard \texttt{CoT} which commits to one line of reasoning).
In effect, \texttt{ToT} is an iterative prompting procedure where the model generates thoughts, evaluates them, and refines its approach, mimicking how a human might mentally map out various ways to solve a problem. 

\texttt{ToT} is especially useful for complex problems like puzzles, planning tasks, or games where multiple steps and strategic exploration are needed and outperforms simpler \texttt{CoT} methods by systematically searching through the solution space. 
It provides a flexible framework – one can plug in various generation strategies (e.g. sampling vs. prompting) and search algorithms \texttt{(BFS, DFS, A*, MCTS)} depending on the task.
Although more computationally heavy, \texttt{ToT} shows that allocating extra “thinking time” (compute) to explore alternatives can yield significantly better reasoning and planning performance. 
It has spawned follow-up research aiming to improve or utilize it for better reasoning e.g., multi-agent systems have been combined with \texttt{ToT}: different \llm ``agents'' generate thoughts in parallel and a validator agent prunes incorrect branches, leading to improved accuracy over the single-agent \texttt{ToT} \cite{haji2024improving}. 
\insightbox{Inference-time computation for \llms can outperform scaling model parameters, especially for challenging reasoning tasks like math problems.}
\subsection{Graph of Thoughts}
The Graph of Thoughts (\texttt{GoT}) \cite{Besta_2024} framework extends the \texttt{ToT} by allowing more flexible and efficient reasoning processes through graph-based structures rather than strict hierarchical trees. Thought representation differs between the two approaches: in \texttt{ToT}, each step in reasoning is structured as a node in a tree with fixed parent-child relationships, whereas \texttt{GoT} represents thoughts as nodes in a graph, enabling more adaptable dependencies and interconnections. 

In terms of thought expansion strategies, \texttt{ToT} follows a traditional approach where multiple thought candidates are generated at each step, explored using tree-based search strategies, and pruned based on heuristics before selecting the most optimal path. In contrast, \texttt{GoT} incorporates graph-based thought expansion, allowing thoughts to interconnect dynamically. This enables three key transformations: aggregation (merging multiple solutions into a unified answer), refinement (iteratively improving thoughts over time), and generation (producing diverse candidates). Instead of navigating through a rigid hierarchy, \texttt{GoT} prioritizes thoughts using a volume metric and explores paths optimally, reducing unnecessary computations.

A critical limitation of \texttt{ToT} is its restricted backtracking—once a branch is discarded, it is not reconsidered. \texttt{GoT} overcomes this by allowing iterative refinement, where previous thoughts can be revisited, modified, and improved upon. This iterative nature is particularly useful in complex reasoning tasks where initial thoughts may require adjustments. Moreover, computational efficiency in \texttt{GoT} is significantly improved by reducing redundant calculations through the merging of partial solutions. 


\insightbox{\texttt{GoT} enhances problem-solving efficiency and adaptability, making it superior to \texttt{ToT} for tasks requiring complex reasoning.}
\subsection{Confidence-based Sampling}
In confidence-based sampling, the language model generates multiple candidate solutions or reasoning paths and then prioritizes or selects among them based on the model’s own confidence in each outcome \cite{wilson2024llm}. This can happen in two ways: (a) Selection: Generate N outputs and pick the one with the highest log probability (i.e., the model’s most confident output). This is essentially best-of-N by probability – the model chooses the answer it thinks is most likely correct. (b) Guided exploration: When exploring a reasoning tree or multi-step solution, use the model’s token probabilities to decide which branch to expand (higher confidence branches are explored first). In other words, the model’s probability estimates act as a heuristic guiding the search through solution space \cite{hendrycks2016baseline}.
Compared to pure random sampling, confidence-based methods bias the process toward what the model believes is right, potentially reducing wasted exploration on low-likelihood (and often incorrect) paths.

Confidence-based strategies have been incorporated at inference time e.g., a tree-based search for \llm generation \cite{wilson2024llm} assigns each possible completion (leaf) a confidence score. 
The algorithm samples leaves in proportion to these confidence scores to decide which paths to extend~\cite{xie-etal-2020-worldtree}.
Similarly, some reasoning approaches use the model’s estimated likelihood of an answer to decide when to halt or whether to ask a follow-up question – essentially if the model’s confidence is low, it might trigger further reasoning (a form of self-reflection). Confidence-based selection is also used in ensemble settings: e.g., an \llm may generate multiple answers and a secondary model evaluates the confidence of each answer being correct, picking the answer with the highest confidence. This was explored in tasks like medical Q\&A, where an \llm gave an answer and a confidence score, and only high confidence answers were trusted or returned \cite{portillowightman2023strength}.
\subsection{Search Against Verifiers}\label{ORM2}
This verification approach \cite{qi2024verifierq} in  \llms enhances answer quality by generating multiple candidate responses and selecting the best one using automated verification systems. This approach shifts focus from increasing pre-training compute to optimizing test-time compute, allowing models to ``think longer'' during inference through structured reasoning steps or iterative refinement. 
The method involves two main steps:\\
\textbf{Generation}: The model (or ``proposer'' produces multiple answers or reasoning paths, often using methods like high-temperature sampling or diverse decoding.\\
\textbf{Verification}: A verifier (e.g., a reward model) evaluates these candidates based on predefined criteria, such as correctness, coherence, or alignment with desired processes.
Verifiers are categorized based on their evaluation focus:
\begin{enumerate}
    \item \textbf{Outcome Reward Models (\texttt{ORM})}: Judge only the final answer (e.g., correctness of a math solution).
    \item \textbf{Process Reward Models (\texttt{PRM})}: Evaluate the reasoning steps (e.g., logical coherence in a thought chain), providing granular feedback to prune invalid paths.
\end{enumerate}
Several techniques fall under this paradigm, enhancing verification-based optimization. Best-of-N Sampling involves generating multiple answers and ranking them via a verifier (\texttt{ORM}/\texttt{PRM}), selecting the highest-scoring one, making it a simple yet effective approach for improving answer correctness. Beam Search with \texttt{PRM} tracks top-scoring reasoning paths (beams) and prunes low-quality steps early, similar to Tree of Thought approaches, balancing breadth and depth in reasoning path exploration. Monte Carlo Tree Search balances exploration and exploitation by expanding promising reasoning branches, simulating rollouts, and backpropagating scores, providing an optimal trade-off between search depth and verification confidence. Majority Voting (Self-Consistency) aggregates answers from multiple samples and selects the most frequent one, avoiding explicit verifiers, which works well in settings where consistency across multiple responses indicates correctness.
\insightbox{\texttt{ORM} is suitable for tasks where correctness is binary (right/wrong) and can be easily assessed.}
\insightbox{\texttt{PRM} is useful in multi-step reasoning, ensuring intermediate steps follows logical progression.}

\subsection{Self-Improvement via Refinements}
This approach refers to the ability of \llms to enhance their outputs through self-evaluation and revision iteratively. This process enables models to refine their responses dynamically during inference rather than relying solely on pre-trained weights.
One notable method is \textbf{Self-Refinement} \cite{madaan2023selfrefineiterativerefinementselffeedback}, where an \llm generates an initial response, critiques it, and then refines the output based on its self-generated feedback. This iterative process continues until the model achieves a satisfactory result. Such techniques have been shown to improve performance on various tasks, including mathematical reasoning and code generation. This process follows these key steps:
a) \textbf{Initial Generation}: The model produces an answer or reasoning path. b) \textbf{Self-Critique}: The model reviews its own response and identifies errors, inconsistencies, or areas for improvement. c) \textbf{Refinement}: The model adjusts its response based on the critique and generates an improved version. d) \textbf{Iteration}: The process repeats until the output meets a predefined quality threshold or stops improving.

Another approach is called \textbf{Self-Polish} \cite{xi2023rise}, where the model progressively refines given problems to make them more comprehensible and solvable. By rephrasing or restructuring problems, the model enhances its understanding and provides more accurate solutions.
Self-Polish involves progressive refinement of problem statements to make them more comprehensible and solvable. The model
first rephrases or restructures the problem for better clarity, then breaks down complex queries into simpler sub-problems and refines ambiguous inputs to ensure precise understanding.
By restructuring problems before solving them, the model improves its comprehension and generates more accurate solutions.
\insightbox{Self-improvement methodologies represent a paradigm shift in \llm optimization, emphasizing active reasoning and internal feedback over static pre-training. By iterating on their own responses, models achieve greater consistency and accuracy across a wide range of applications.}
\subsection{Monte Carlo Tree Search}
\texttt{MCTS}~\cite{coulom2006efficient} is based on the application of Monte Carlo simulations to game-tree search. 
It rose to prominence with successes in games, notably, it powered AlphaGo \cite{chen2016evolution} in 2016 by searching possible moves guided by policy and value networks. 
This, as well as the application to other board and video games, demonstrates the power of \texttt{MCTS} for sequential decision-making under uncertainty.

\texttt{MCTS} is a stochastic search algorithm that builds a decision tree by performing many random simulations. It is best known for finding good moves in game states, but it can be applied to any problem where we can simulate outcomes. The algorithm iteratively: (a) Selects a path from the root according to a heuristic (like \texttt{UCT} \cite{kocsis2006bandit}, which picks nodes with a high upper-confidence bound), (b) Expands a new node (a previously unvisited state) from the end of that path, (c) Simulates a random rollout from that new state to get an outcome (e.g., win or loss in a game, or some reward), and (d) Backpropagates the result up the tree to update the values of nodes and inform future selections. Repeating these simulations thousands of times concentrates the search on the most promising branches of the tree. In essence, \texttt{MCTS} uses random sampling to evaluate the potential of different action sequences, gradually biasing the search towards those with better average outcomes. In \llm reasoning, we can treat the generation of text as a decision process and use \texttt to explore different continuations. For example, at a given question (root), each possible next reasoning step or answer is an action; a simulation could mean letting the \llm continue to a final answer (perhaps with some randomness), and a reward could be whether the answer is correct. By doing this repeatedly, \texttt{MCTS} can identify which chain of thoughts or answers has the highest empirical success rate. The appeal of \texttt{MCTS} for reasoning is that it can handle large search spaces by sampling intelligently rather than exhaustively, and it naturally incorporates uncertainty and exploration.
\insightbox{Train verifiers to score intermediate steps (via Monte Carlo rollouts) instead of just final answers.}
Recent efforts have integrated \texttt{MCTS} with \llms to tackle complex reasoning and decision-making tasks. One example is using \texttt{MCTS} for query planning: Monte Carlo Thought Search \cite{sprueill2023monte}, where an \llm is guided to ask a series of sub-questions to find an answer. Jay et al. \cite{sprueill2023monte} used an \texttt{MCTS}-based algorithm called `Monte Carlo Reasoner' that treats the \llm as an environment: each node is a prompt (state) and each edge is an action (e.g., a particular question to ask or step to take), and random rollouts are used to evaluate outcomes. 
This approach allowed the system to efficiently explore a space of possible reasoning paths and pick a high-reward answer path, outperforming naive sampling in a scientific Q\&A task.
Similarly, \texttt{MCTS} has been applied to code generation with \llms \cite{delorenzo2024make} – the algorithm explores different code paths (using the model to propose code completions and etest them) to find a correct solution. 
Another line of work ensembles multiple \llms with \texttt{MCTS}, treating each model’s output as a branch and using a reward model to simulate outcomes \cite{park2024ensembling}. 
Early results show that \texttt{MCTS}-based reasoning can solve problems that single-pass or greedy methods often miss, although with more compute \cite{xie2024monte}. 
The downside is that \texttt{MCTS} can be significantly slower than straightforward sampling or beam search, which recent research is addressing by improving efficiency (e.g., by state merging \cite{tian2024toward}). 
In general, \texttt{MCTS} brings the strength of planning algorithms to \llm inference and enables an \llm to 'look ahead' through simulated rollouts and make more informed reasoning choices, much like it has done for AI in gameplay.
\insightbox{Test-time compute is not a 1-to-1 replacement for pretraining but, offers a viable alternative in many cases.}
\subsection{Chain-of-Action-Thought reasoning}
\llms excel in reasoning tasks but rely heavily on external guidance (e.g., verifiers) or extensive sampling at inference time. Existing methods like \texttt{CoT} \cite{wei2022chain} lack mechanisms for self-correction and adaptive exploration, limiting their autonomy and generalization. Satori \cite{shen2025satori} introduced a two-stage training paradigm, which works by initially tuning the model’s output format and then enhancing its reasoning capabilities through self-improvement. In Stage 1 (Format Tuning), the model is exposed to a large set of 10K synthetic trajectories generated by a multi-agent framework comprising a generator, a critic, and a reward model. This supervised fine-tuning helps the model to produce outputs in specific reasoning format using meta-action tokens, although it may still have difficulty generalizing beyond these examples. In Stage 2 (Self-Improvement via \rl), the model employs \ppo with a Restart and Explore strategy \cite{shen2025satori}, which allows it to restart from intermediate steps, whether they were correct or not, to refine its reasoning process. The model receives rewards based on a combination of rule-based correctness, reflection bonuses, and preference-based Outcome Reward Model feedback explained in \S~\ref{ORM2}, thereby incentivizing the allocation of more computational resources to tougher problems and enabling extended reasoning during testing for complex tasks.

Multi-agent frameworks and advanced fine-tuning strategies are increasingly being explored to enhance reasoning in \llms. Multi-Agent \llm Training (\texttt{MALT})~\cite{motwani2025maltimprovingreasoningmultiagent} introduces a structured approach where generation, verification, and refinement steps are distributed across specialized agents, allowing for iterative self-correction and improved reasoning chains. Similarly, optimizing preference alignment remains a crucial challenge in ensuring both safety and helpfulness in \llms~\cite{anwar2024foundational}. Approaches like Bi-Factorial Preference Optimization (\texttt{BFPO})~\cite{zhang2024bi} reframe \rlhf objectives into a single supervised learning task, reducing human intervention while maintaining robust alignment. Beyond text-based reasoning, multimodal approaches like Multimodal Visualization-of-Thought (\texttt{MVoT})~\cite{li2025imagine} extend \texttt{CoT} prompting by incorporating visual representations, significantly enhancing performance in spatial reasoning tasks. These advancements highlight the growing need for structured multi-agent collaboration, safety-aware optimization, and multimodal reasoning to address fundamental limitations in \llm reasoning~\cite{nowak2024representational, li2024chain, merrill2023expressive}.

\subsection{Pretraining vs. Test-Time Scaling}
Pretraining and \texttt{TTS} are two distinct strategies for improving \llm performance, each with different tradeoffs in computational cost and effectiveness. Pretraining involves scaling model parameters or increasing training data to enhance capabilities, requiring substantial upfront computational investment~\cite{yang2019xlnet}. In contrast, \texttt{TTS} optimizes inference-time compute (such as iterative refinements, search-based decoding, or adaptive sampling), allowing performance improvements without modifying the base model.

From a performance vs. cost perspective, \texttt{TTS} achieves results comparable to a model 14× larger on easy to intermediate tasks (e.g., MATH benchmarks), while reducing inference costs by 4× fewer FLOPs in compute-intensive scenarios~\cite{snellscaling2025}. However, pretraining remains superior for the hardest tasks or when inference compute constraints are high, as larger pretrained models inherently encode deeper reasoning capabilities.
\insightbox{A smaller model with test-time compute can outperform a 14× larger model on easy/intermediate questions, when inference tokens (Y) are limited (e.g., self-improvement settings).}
In terms of use cases, \texttt{TTS} is useful for scenarios with flexible inference budget or when base models already exhibit reasonable competence in the task. Conversely, pretraining is essential for tasks requiring fundamentally new capabilities (e.g., reasoning on novel domains) where inference-time optimizations alone may not suffice.

There are notable tradeoffs between the two approaches. \texttt{TTS} reduces upfront training costs, making it attractive for flexible, on-the-go optimization, but requires dynamic compute allocation at inference. Pretraining, on the other hand, incurs high initial costs but guarantees consistent performance without additional runtime overhead, making it ideal for large-scale \texttt{API} deployments or latency-sensitive applications. Overall, \texttt{TTS} and pretraining are complementary in nature. Future \llm systems may adopt a hybrid approach, where smaller base models are pretrained with essential knowledge, while \texttt{TTS} dynamically enhances responses through adaptive, on-demand computation. This synergy enables more cost-effective and efficient large-scale model deployment. 

\insightbox{Choose pretraining for foundational capabilities and test-time scaling for accurate context-aware refinement.}

%% file: figures/PPO_comp_fig.tex
\begin{figure}
    \centering
    \includegraphics[width=\columnwidth]{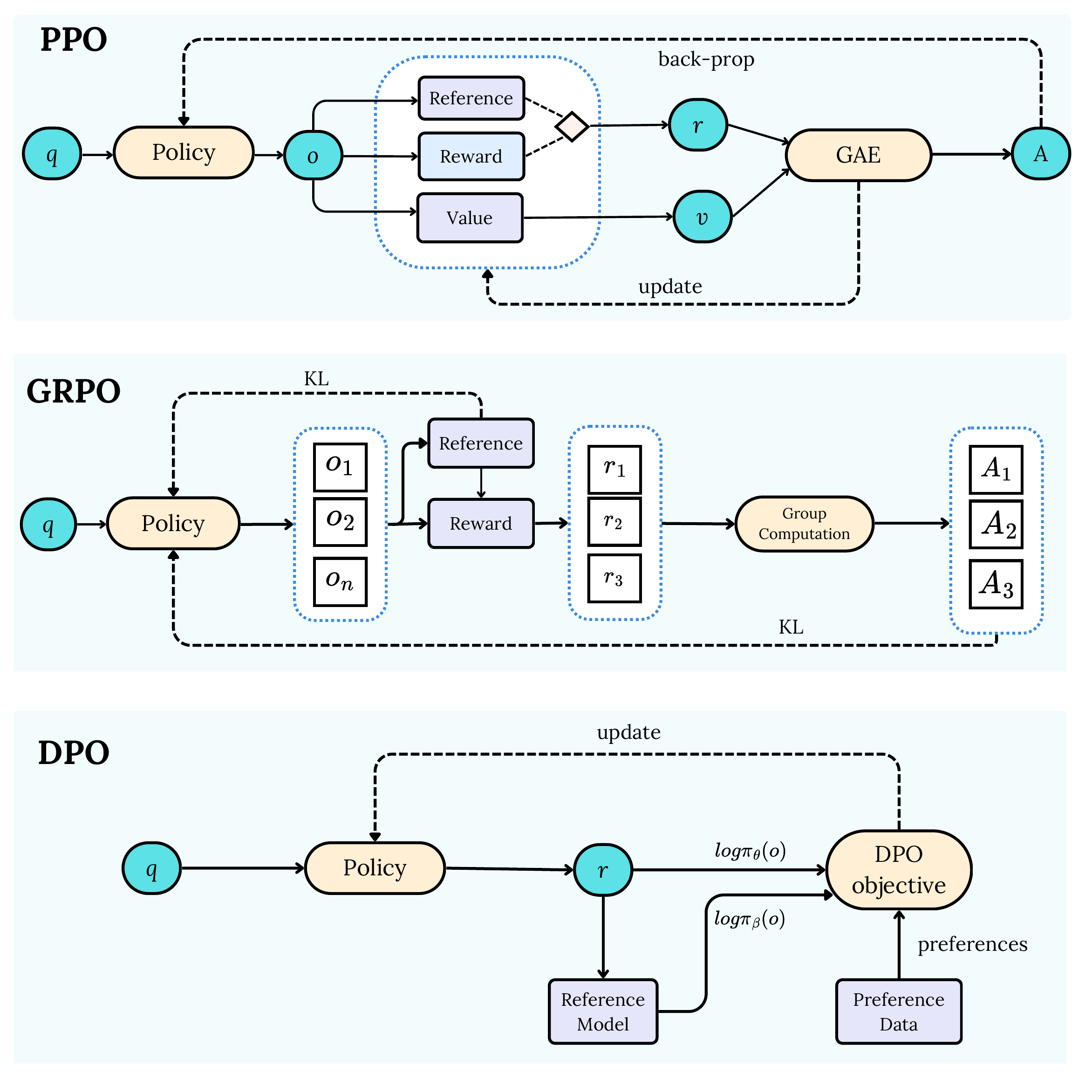}
    \caption{Comparison of \ppo~\cite{schulman2017proximal}, \grpo~\cite{shao2024deepseekmath}, and \dpo~\cite{guo2024direct}. We highlight policy models, reference models, rewards and optimization flows with corresponding loss functions.}
    \label{fig:enter-label}
\end{figure}

%% file: tables/methods.tex
\begin{table*}[ht]
\centering
\scriptsize  
\setlength{\tabcolsep}{4pt} 
\resizebox{\textwidth}{!}{
\begin{tabular}{@{}lll>{\raggedright\arraybackslash}p{10cm}@{}}
\toprule

\textbf{Model} & \textbf{Category}& \textbf{Source} & \textbf{Description} \\
\midrule
\rowcolor{gray!10}
\multicolumn{4}{l}{\textbf{1. Parameter-Efficient Fine-Tuning \& Model Compression}} \\
\midrule
\textbf{LoRA}~\cite{hu2021lora} & Low-Rank Adaptation & \href{https://github.com/microsoft/LoRA}{Link} & Injects trainable low-rank adapters for efficient fine-tuning. \\
\textbf{QLoRA}~\cite{dettmers2023qlora} & Quantized Adaptation & \href{https://github.com/artidoro/qlora}{Link} & Combines 4-bit quantization with LoRA to enable fine-tuning on consumer GPUs \\
\textbf{GPTQ}~\cite{frantar-gptq} & Post-Training Quantization & \href{https://github.com/IST-DASLab/gptq}{Link} & Optimal 4-bit quantization method for GPT-style models with minimal loss \\
\textbf{SparseGPT}~\cite{frantar-sparsegpt} & Pruning & \href{https://github.com/IST-DASLab/sparsegpt}{Link} & One-shot pruning that preserves model quality with compensation. \\
\textbf{PEFT (HF)}~\cite{peft} & Unified Fine-Tuning & \href{https://github.com/huggingface/peft}{Link} & Library integrating LoRA, prefix tuning, and other parameter-efficient methods \\
\textbf{BitsAndBytes}~\cite{dettmers2022llmint8} & Low-Precision Training & \href{https://github.com/TimDettmers/bitsandbytes}{Link} & Enables 8-bit optimizers and 4-bit quantization for memory-efficient training \\
\textbf{AdaLoRA}~\cite{zhang2023adaptive} & Adaptive Adaptation & \href{https://github.com/QingruZhang/AdaLoRA}{Link} & Dynamically allocates parameter budget between layers during fine-tuning \\
\textbf{P-Tuning v2}~\cite{DBLP:journals/corr/abs-2110-07602} & Prompt Optimization & \href{https://github.com/THUDM/P-tuning-v2}{Link} & Learns continuous prompt embeddings through deep prompt tuning \\
\midrule
\rowcolor{gray!10}
\multicolumn{4}{l}{\textbf{2. Data Management \& Preprocessing}} \\
\midrule

\textbf{HF Datasets}~\cite{lhoest-etal-2021-datasets} & Data Processing & \href{https://github.com/huggingface/datasets}{Link} & Unified API for 30k+ datasets with streaming, versioning, and preprocessing \\
\textbf{WebDataset}~\cite{webdataset} & Data Streaming & \href{https://github.com/webdataset/webdataset}{Link} & Efficient tar-based sharding format for petascale distributed training \\
\textbf{DVC}~\cite{dvc} & Data Versioning & \href{https://github.com/iterative/dvc}{Link} & Git-like version control for datasets and machine learning pipelines \\
\textbf{Apache Arrow}~\cite{apache} & Memory Format & \href{https://github.com/apache/arrow}{Link} & Language-agnostic columnar memory format for zero-copy data access \\
\textbf{Zstandard}~\cite{zstandard} & Compression & \href{https://github.com/facebook/zstd}{Link} & High-speed compression algorithm for training data storage/transfer \\
\textbf{Cleanlab}~\cite{cleanlab} & Data Quality & \href{https://github.com/cleanlab/cleanlab}{Link} & Automatic detection of label errors and outliers in training datasets \\
\midrule
\rowcolor{gray!10}
\multicolumn{4}{l}{\textbf{3. Distributed Training \& Optimization}} \\
\midrule

\textbf{DeepSpeed}~\cite{aminabadi2022deepspeedinferenceenablingefficient} & Training Optimization & \href{https://github.com/microsoft/DeepSpeed}{Link} & ZeRO parallelism, 3D parallelism, and memory optimizations for giant models \\
\textbf{Megatron-LM}~\cite{shoeybi2020megatronlmtrainingmultibillionparameter} & Model Parallelism & \href{https://github.com/NVIDIA/Megatron-LM}{Link} & NVIDIA's optimized framework for large transformer model training \\
\textbf{Colossal-AI}~\cite{li2023colossalaiunifieddeeplearning} & Heterogeneous Training & \href{https://github.com/hpcaitech/ColossalAI}{Link} & Unified system supporting multiple parallelization strategies \\
\textbf{Horovod}~\cite{sergeev2018horovodfasteasydistributed} & Distributed Training & \href{https://github.com/horovod/horovod}{Link} & MPI-inspired framework for multi-GPU/multi-node synchronization \\
\textbf{Ray}~\cite{moritz2018raydistributedframeworkemerging} & Distributed Computing & \href{https://github.com/ray-project/ray}{Link} & Universal framework for distributed Python applications at scale \\
\midrule
\rowcolor{gray!10}
\multicolumn{4}{l}{\textbf{4. Efficient Inference \& Deployment}} \\
\midrule

\textbf{vLLM}~\cite{kwon2023efficientmemorymanagementlarge} & Serving Optimization & \href{https://github.com/vllm-project/vllm}{Link} & Paged attention implementation for high-throughput LLM serving \\
\textbf{TensorRT}~\cite{10074837} & GPU Optimization & \href{https://github.com/NVIDIA/TensorRT}{Link} & NVIDIA's inference optimizer with kernel fusion and quantization support \\
\textbf{Triton}~\cite{tillet2019triton} & Serving Framework & \href{https://github.com/triton-inference-server/server}{Link} & Production-grade serving with concurrent model execution support \\
\textbf{ONNX}~\cite{onnx} & Cross-Platform & \href{https://github.com/onnx/onnx}{Link} & Unified inference engine with hardware-specific optimizations \\
\textbf{OpenVINO}~\cite{openvino2025} & Intel Optimization & \href{https://github.com/openvinotoolkit/openvino}{Link} & Runtime for Intel CPUs/iGPUs with pruning/quantization support \\
\textbf{XNNPACK}~\cite{dukhan2019indirectconvolutionalgorithm} & Mobile Inference & \href{https://github.com/google/XNNPACK}{Link} & Highly optimized floating-point kernels for ARM CPUs \\
\textbf{Groq}~\cite{groq2025} & AI Accelerator & \href{https://github.com/groq}{Link} & Deterministic low-latency inference via custom tensor streaming processor \\
\midrule
\rowcolor{gray!10}
\multicolumn{4}{l}{\textbf{5. Integrated Development Ecosystems}} \\
\midrule

\textbf{HF Ecosystem}~\cite{castaño2024analyzingevolutionmaintenanceml} & Full Stack & \href{https://github.com/huggingface}{Link} & Transformers + Datasets + Accelerate + Inference Endpoints \\
\textbf{DeepSpeed}~\cite{aminabadi2022deepspeedinferenceenablingefficient} & Training/Inference & \href{https://github.com/microsoft/DeepSpeed}{Link} & Microsoft's end-to-end solution for billion-parameter models \\
\textbf{PyTorch}~\cite{paszke2017automatic} & Unified Framework & \href{https://github.com/pytorch/pytorch}{Link} & Native LLM support via torch.compile and scaled dot-product attention \\
\textbf{LLM Reasoners}~\cite{hao2024llm} & Advanced Reasoning & \href{https://github.com/pytorch/pytorch}{Link} & Enhances \llm reasoning capabilities using advanced search algorithms. \\  
\bottomrule
\end{tabular}
}
\caption{Comprehensive Overview of Methods and Frameworks employed in Modern LLMs}
\label{tab:llm_tools}
\end{table*}

%% file: content/datasets.tex
\section{Benchmarks for \llm Post-training Evaluation}
\label{sec:benchmarks}
To evaluate the success of \llm post-training phases, a diverse set of benchmarks have been proposed covering multiple domains: reasoning tasks, alignment, multilinguality, general comprehension, and dialogue and search tasks. A well-structured evaluation framework ensures a comprehensive understanding of an \llm strengths, and limitations across various tasks. These benchmarks play a crucial role in \llm post-processing stages, where models undergo fine-tuning, calibration, alignment, and optimization to improve response accuracy, robustness, and ethical compliance. Next, we explain the main benchmark gorups.
Table~\ref{tab:reasoning_datasets_all} provides an overview of key datasets categorized under these benchmark groups. 

\input{tables/datasets_test}

\noindent\textbf{Reasoning Benchmarks.}
These benchmarks assess \llms on their ability to perform logical, mathematical, and scientific reasoning. Mathematical reasoning datasets like MATH~\cite{2019arXiv}, GSM8K~\cite{cobbe2021gsm8k}, and MetaMathQA~\cite{yu2023metamath} test models on problem-solving, multi-step arithmetic, and theorem-based problem formulations. 
Scientific and multimodal reasoning benchmarks such as WorldTree V2~\cite{xie-etal-2020-worldtree} and MMMU~\cite{yue2023mmmu} evaluate knowledge in physics, chemistry, and multimodal understanding, which are crucial for fact-checking and verification processes in LLM-generated responses. 
Additionally, datasets like PangeaBench~\cite{liu-etal-2021-visually} extend reasoning tasks into multilingual and cultural domains, enabling models to refine cross-lingual reasoning. These benchmarks help determine how well models can process structured knowledge and apply logical deductions.

\noindent\textbf{RL Alignment Benchmarks.}
\rl alignment benchmarks are central to \llm alignment and post-training optimization. They refine response generation, ethical constraints, and user-aligned outputs through \rlhf. Datasets such as HelpSteer~\cite{wang2023helpsteer} and UltraFeedback~\cite{cui2023ultrafeedback} evaluate models based on multi-attribute scoring and alignment with user instructions. Anthropic's HH-RLHF~\cite{anthropic2025claude37} explores how well models learn human preference optimization through reinforcement learning with human feedback. D4RL~\cite{fu2020d4rl} and Meta-World~\cite{schmied2024learning} focus on robotic control and offline RL, which have implications for autonomous model decision-making. MineRL~\cite{guss2019minerllargescaledatasetminecraft} extends \rl testing into complex environments such as Minecraft-based interactions, useful for training \llms in adaptive decision-making settings.

\noindent\textbf{Multilingual Evaluation.}
Multilingual benchmarks are essential for \llm post-processing in cross-lingual generalization, translation adaptation, and fine-tuning for low-resource languages. CulturaX~\cite{nguyen-etal-2024-culturax} and PangeaIns~\cite{yue2024pangeafullyopenmultilingual} evaluate tokenization, translation, and instruction-following in over 150 languages, ensuring fairness and diversity in model outputs. TydiQA~\cite{DBLP:journals/corr/abs-1810-04805} and MM-Eval~\cite{son2024mmevalmultilingualmetaevaluationbenchmark} target bilingual and task-oriented multilingual evaluation, enabling improvements in \llm fine-tuning. These datasets ensure that \llms are not just English-centric but optimized for multilingual adaptability.

\noindent\textbf{General Comprehension Benchmarks.}
General comprehension benchmarks contribute to model fine-tuning, response coherence, and preference optimization. Datasets such as Chatbot Arena~\cite{zheng2023judging}, MTBench~\cite{zheng2023judging}, and RewardBench~\cite{lambert2024rewardbench} test user preference modeling and conversational fluency, crucial for \llm response ranking and re-ranking methods. BigBench~\cite{https://doi.org/10.48550/arxiv.2206.04615} evaluates broad multi-domain comprehension, while MMLU~\cite{hendryckstest2021, hendrycks2021ethics} measures correctness and informativeness. These datasets help in refining \llm fluency, factual correctness, and open-ended response generation.

\noindent\textbf{Dialogue and Search Benchmarks.}
Dialogue and search benchmarks play a key role in optimizing \llm retrieval-based responses, multi-turn coherence, and information retrieval accuracy.
Datasets such as ConvAI2~\cite{DBLP:journals/corr/abs-1902-00098} and MultiWOZ~\cite{eric-etal-2020-multiwoz} evaluate multi-turn conversational models, essential for dialogue history tracking and adaptive response fine-tuning. For search relevance assessment, BEIR~\cite{thakur2021beir} provides large-scale human-annotated judgments for retrieval fine-tuning, ensuring \llms generate and rank responses effectively. TREC DL21/22~\cite{li-roth-2002-learning, hovy-etal-2001-toward} contributes to document relevance ranking and fact retrieval.

%% file: tables/datasets_test.tex
\begin{table}[tbp]
\centering\setlength{\tabcolsep}{1pt}
\caption{\small Comprehensive Overview of Reasoning, RL Alignment, and Multilingual Datasets. Here, pointwise and pairwise refer to different methods of evaluating model performance across various tasks.}

\label{tab:reasoning_datasets_all}
\begin{adjustbox}{width=\columnwidth}
\begin{tabular}{lcccc}
\toprule
\rowcolor{gray!20}
\textbf{Datasets} & \textbf{Domain} & \textbf{Type} & \textbf{\#Samples} & \textbf{Evaluation Criteria} \\
\midrule

\multicolumn{5}{c}{\textbf{Reasoning Benchmarks}} \\
\midrule
\textbf{MATH} ~\cite{2019arXiv}                     & Math Reasoning        & Pointwise             & 7,500  & Step-by-step solutions  \\
\rowcolor{gray!10}
\textbf{GSM8K}  ~\cite{cobbe2021gsm8k}                   & Math Reasoning        & Pointwise             & 8.5K   & Multi-step reasoning    \\
\textbf{MetaMathQA}~\cite{yu2023metamath}                 & Math Reasoning        & Pointwise             & 40K+   & Self-verification, FOBAR \\
\rowcolor{gray!10}
\textbf{WorldTree V2}  ~\cite{xie-etal-2020-worldtree}             & Science QA           & Pointwise             & 1,680  & Multi-hop explanations  \\
\textbf{PangeaBench}  ~\cite{liu-etal-2021-visually}              & Multimodal Reasoning  & Pairwise              & 47 Langs.  & Cultural understanding  \\
\rowcolor{gray!10}
\textbf{MMMU}    ~\cite{yue2023mmmu}                   & Science/Math  & Pointwise           & College-Level  & Physics, Chemistry, Bilingual  \\
\textbf{TruthfulQA}  ~\cite{lin2021truthfulqa}               & QA/Reasoning         & Pointwise             & N/A    & Truthfulness  \\
\rowcolor{gray!10}
\textbf{MathInstruct} ~\cite{yue2023mammoth}              & Math Reasoning       & Pointwise                   & 262K   & Correctness  \\
\rowcolor{gray!10}
\textbf{MMLU} ~\cite{hendryckstest2021, hendrycks2021ethics}                      & Multitask Reasoning  & Pointwise                  & 57 Tasks & Broad knowledge evaluation  \\
\textbf{MMLU-Fairness} ~\cite{hendryckstest2021}             & Fairness/Reasoning   & Pointwise                  & N/A    & Bias/Equity Analysis  \\
\rowcolor{gray!10}
\textbf{DROP}     ~\cite{Dua2019DROP}                  & Reading/Reasoning    & Pointwise                   & 96K    & Discrete reasoning over paragraphs  \\
\textbf{BBH} ~\cite{suzgun2022challenging}                       & Hard Reasoning       & Pairwise                   & N/A    & Complex logical problem-solving  \\
\textbf{VRC-Bench}~\cite{thawakar2025llamavo1}                          & Multimodal Reasoning  & Pairwise             & N/A    & Visual Reasoning and Classification  \\
\midrule
\multicolumn{5}{c}{\textbf{RL Alignment Benchmarks}} \\
\midrule
\textbf{HelpSteer}  ~\cite{wang2023helpsteer}                & RL Alignment         & Pairwise              & 37K+   & Multi-attribute scoring  \\
\rowcolor{gray!10}
\textbf{Anthropic HH-RLHF} ~\cite{anthropic2025claude37}         & RL Alignment         & Pairwise              & 42.5K  & Harmlessness alignment  \\
\textbf{UltraFeedback}  ~\cite{cui2023ultrafeedback}            & RL Alignment         & Pairwise              & 64K    & Instruction-following, Truthfulness  \\
\rowcolor{gray!10}
\textbf{D4RL} ~\cite{fu2020d4rl}                      & RL/Control           & Pointwise                   & N/A    & Offline RL across domains  \\
\textbf{Meta-World}  ~\cite{schmied2024learning}               & RL/Control           & Pointwise                  & N/A    & Multi-task robotic RL  \\
\rowcolor{gray!10}
\textbf{MineRL} ~\cite{guss2019minerllargescaledatasetminecraft}                    & RL/Games            & Pairwise                  & N/A    & Imitation learning, rewards  \\

\midrule
\multicolumn{5}{c}{\textbf{Multilingual Evaluation}} \\
\midrule
\textbf{CulturaX} ~\cite{nguyen-etal-2024-culturax}                  & Multilingual        & Pointwise             & 6.3T   & Deduplication, Quality  \\
\rowcolor{gray!10}
\textbf{PangeaIns} ~\cite{yue2024pangeafullyopenmultilingual}                 & Multilingual        & Pointwise             & 6M     & Multilingual instructions  \\
\textbf{TydiQA}~\cite{DBLP:journals/corr/abs-1810-04805}                     & Multilingual        & Pointwise             & N/A    & Cross-lingual QA  \\
\rowcolor{gray!10}
\textbf{XGLUE} ~\cite{Liang2020XGLUEAN}                     & Multilingual        & Pointwise             & N/A    & Cross-lingual language tasks  \\
\textbf{MM-Eval}~\cite{son2024mmevalmultilingualmetaevaluationbenchmark}                    & Multilingual        & Pairwise              & 4,981  & Task-oriented multilingual QA  \\
\textbf{ALM-Bench}~\cite{son2024mmevalmultilingualmetaevaluationbenchmark}                        & Multilingual QA       & Pointwise             & N/A    & Multilingual Evaluation  \\

\midrule
\multicolumn{5}{c}{\textbf{Dialogue and Search Benchmarks   }} \\
\midrule
\textbf{BigBench}~\cite{https://doi.org/10.48550/arxiv.2206.04615}                   & General Comprehension & Pointwise           & 200+ Tasks  & Broad multi-domain evaluation  \\
\rowcolor{gray!10}
\textbf{Chatbot Arena}~\cite{zheng2023judging}              & Comprehension       & Pairwise              & 33K    & User preference  \\
\textbf{MTBench} ~\cite{zheng2023judging}                   & Comprehension       & Pairwise              & 3K     & Multi-turn conversations  \\
\rowcolor{gray!10}
\textbf{RewardBench}~\cite{lambert2024rewardbench}                & Comprehension       & Pairwise              & 2,998  & User preference  \\

\midrule
\multicolumn{5}{c}{\textbf{General Comprehension Benchmarks}} \\
\midrule
\textbf{ConvAI2} ~\cite{DBLP:journals/corr/abs-1902-00098}                   & Dialogue            & Pointwise             & N/A    & Engagingness, Consistency  \\
\rowcolor{gray!10}
\textbf{MultiWOZ}~\cite{eric-etal-2020-multiwoz}                   & Dialogue            & Pointwise             & N/A    & Task success, Coherence  \\
\textbf{Trec DL21\&22}~\cite{li-roth-2002-learning, hovy-etal-2001-toward}              & Search              & Pointwise             & 1,549/2,673  & Relevance scoring  \\
\rowcolor{gray!10}
\textbf{BEIR}~\cite{thakur2021beir}                      & Search              & Pointwise             & 18 Datasets  & Information retrieval  \\

\midrule
\multicolumn{5}{c}{\textbf{Story \& Recommendation Benchmarks}} \\
\midrule
\textbf{HANNA} ~\cite{chhun2024do}                     & Story               & Pointwise             & 1,056  & Relevance, Coherence, Complexity  \\
\rowcolor{gray!10}
\textbf{StoryER}~\cite{chen2022storyerautomaticstoryevaluation}                    & Story               & Pairwise              & 100K   & User preference-based ranking  \\
\textbf{PKU-SafeRLHF}~\cite{ji2024beavertails}              & Values              & Pairwise              & 83.4K  & Helpfulness, Harmlessness  \\
\rowcolor{gray!10}
\textbf{Cvalue}~\cite{xu2023cvalues}                   & Values              & Pairwise              & 145K   & Safety, Responsibility  \\
\textbf{NaturalInst.}~\cite{naturalinstructions,supernaturalinstructions}       & Instruction Tuning  & Pointwise                  & 1,600+  & Instruction-following evaluation  \\

\bottomrule
\end{tabular}
\end{adjustbox}
\end{table}

%% file: content/directions.tex
\section{Future Directions}
\begin{figure*}[h]
    \centering
    \begin{subfigure}[b]{0.32\textwidth}
        \centering
        \includegraphics[width=\textwidth]{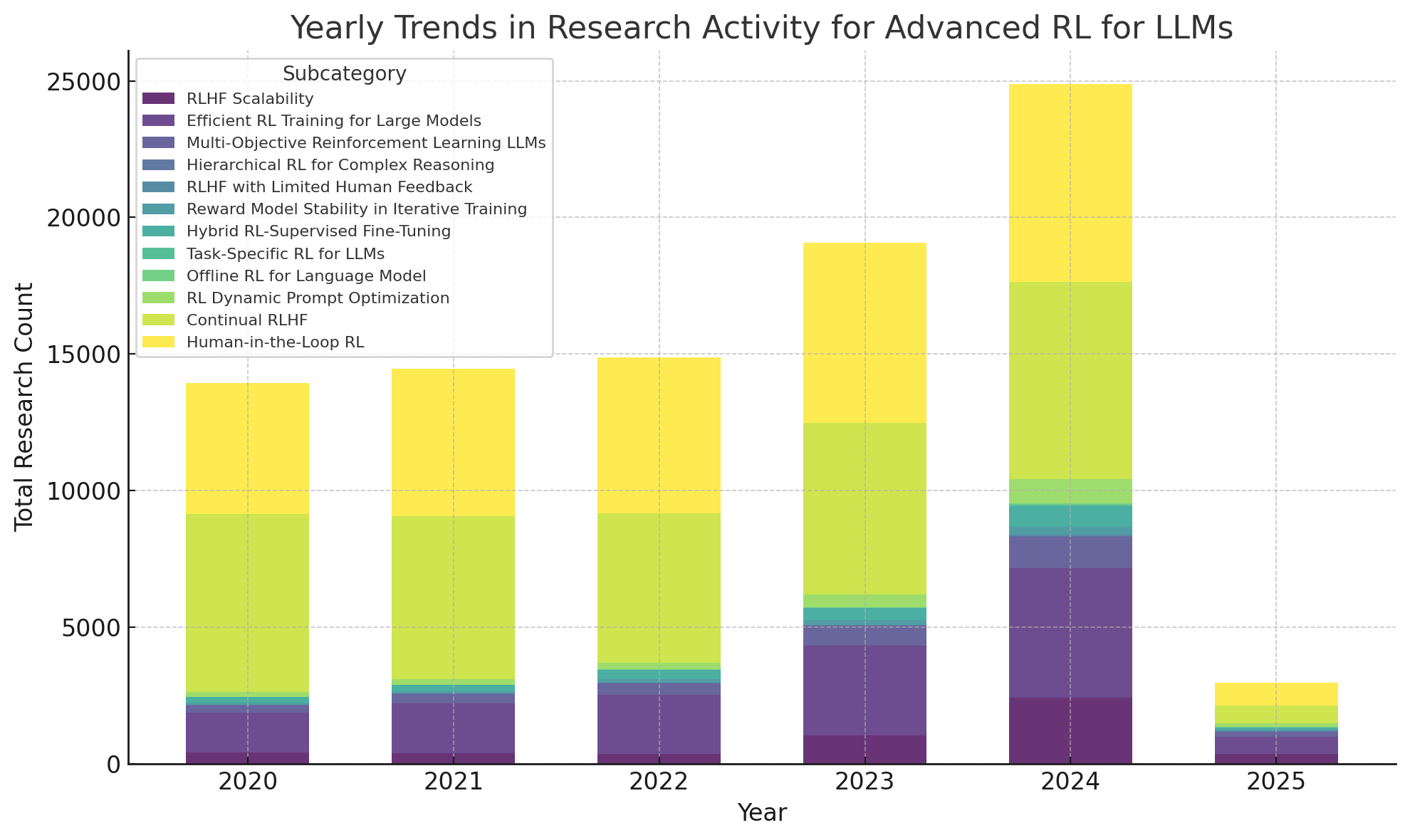}
        \caption{\small Growing trend in RL for LLMs, with a focus on Human-in-the-Loop RL.}
        \label{fig:advanced_rl}
    \end{subfigure}
    \hfill
    \begin{subfigure}[b]{0.33\textwidth}
        \centering
        \includegraphics[width=\textwidth]{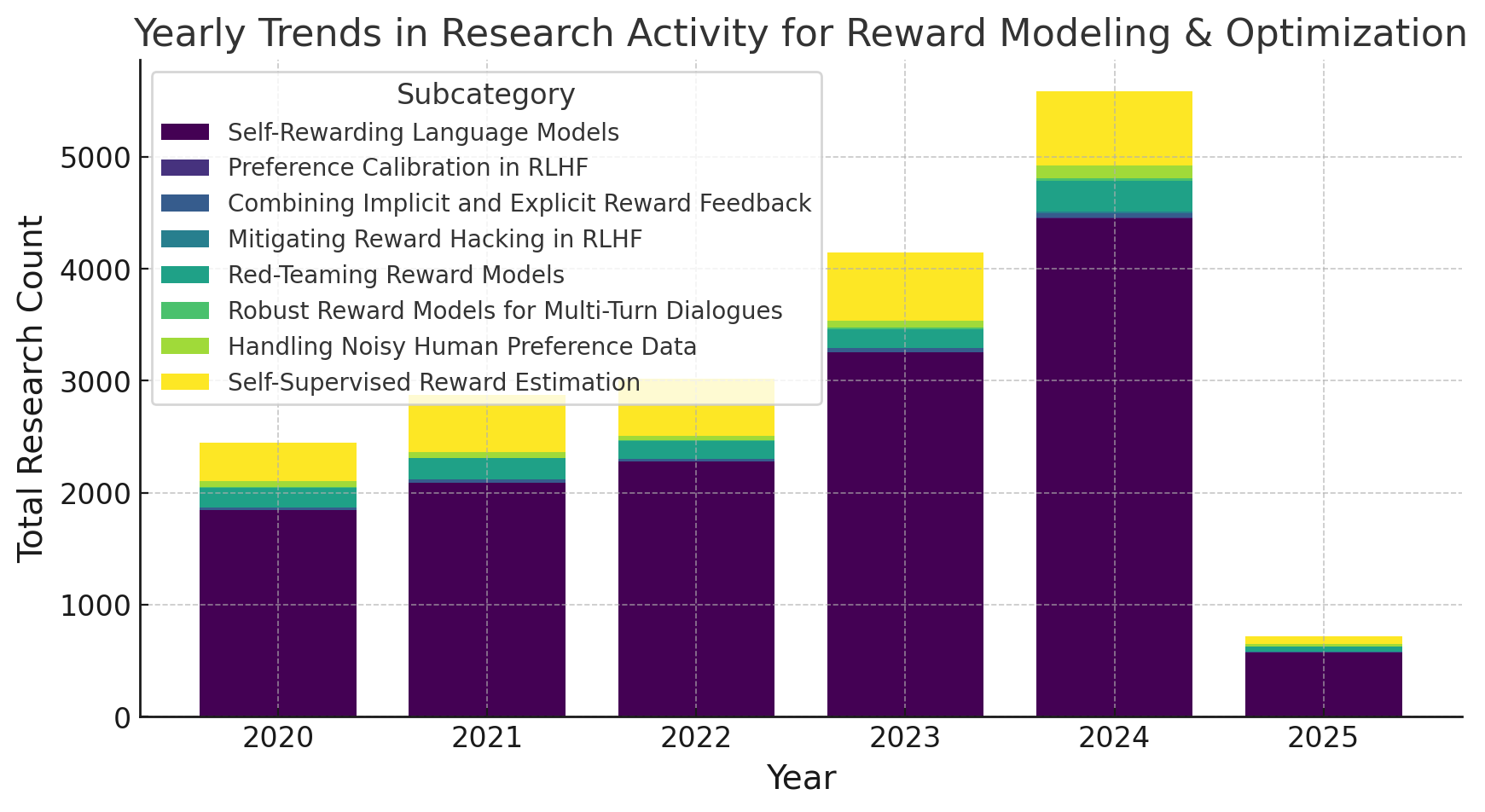}
        \caption{\small Reward modeling trends show RLHF stabilization, with Self-Rewarding Models leading, but Reward Hacking persists.}
        \label{fig:reward_modeling}
    \end{subfigure}
    \hfill
    \begin{subfigure}[b]{0.32\textwidth}
        \centering
        \includegraphics[width=\textwidth]{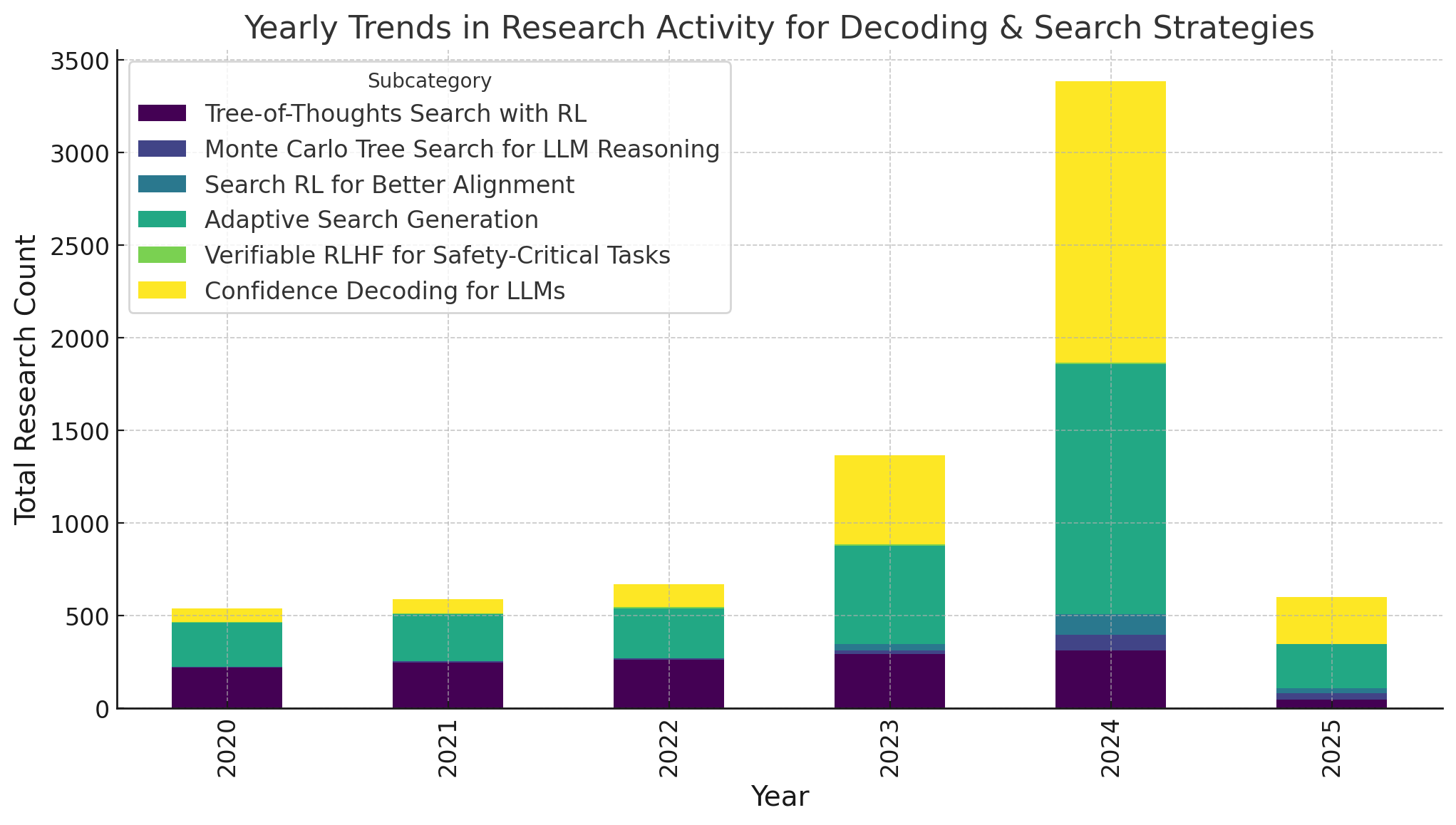}
        \caption{\small Decoding strategies like Tree-of-Thoughts and MCTS are improving LLM reasoning and decision-making.}
        \label{fig:decoding}
    \end{subfigure}
    \begin{subfigure}[b]{0.32\textwidth}
        \centering
        \includegraphics[width=\textwidth]{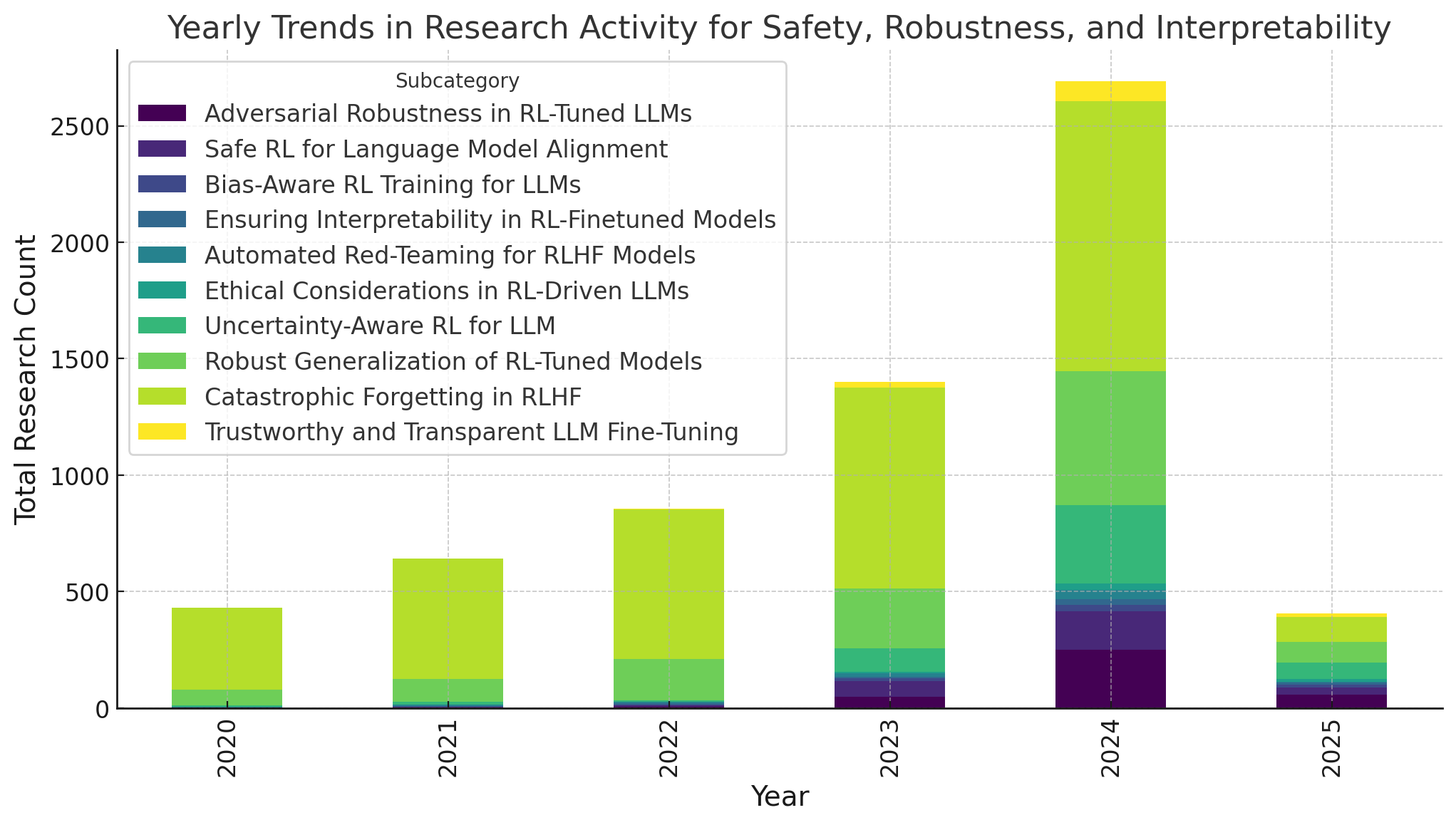}
        \caption{\small Safety and Robustness research is growing, with Uncertainty-Aware \rl ensuring \rlhf model reliability.}
        \label{fig:safety}
    \end{subfigure}
    \hfill
    \begin{subfigure}[b]{0.32\textwidth}
        \centering
        \includegraphics[width=\textwidth]{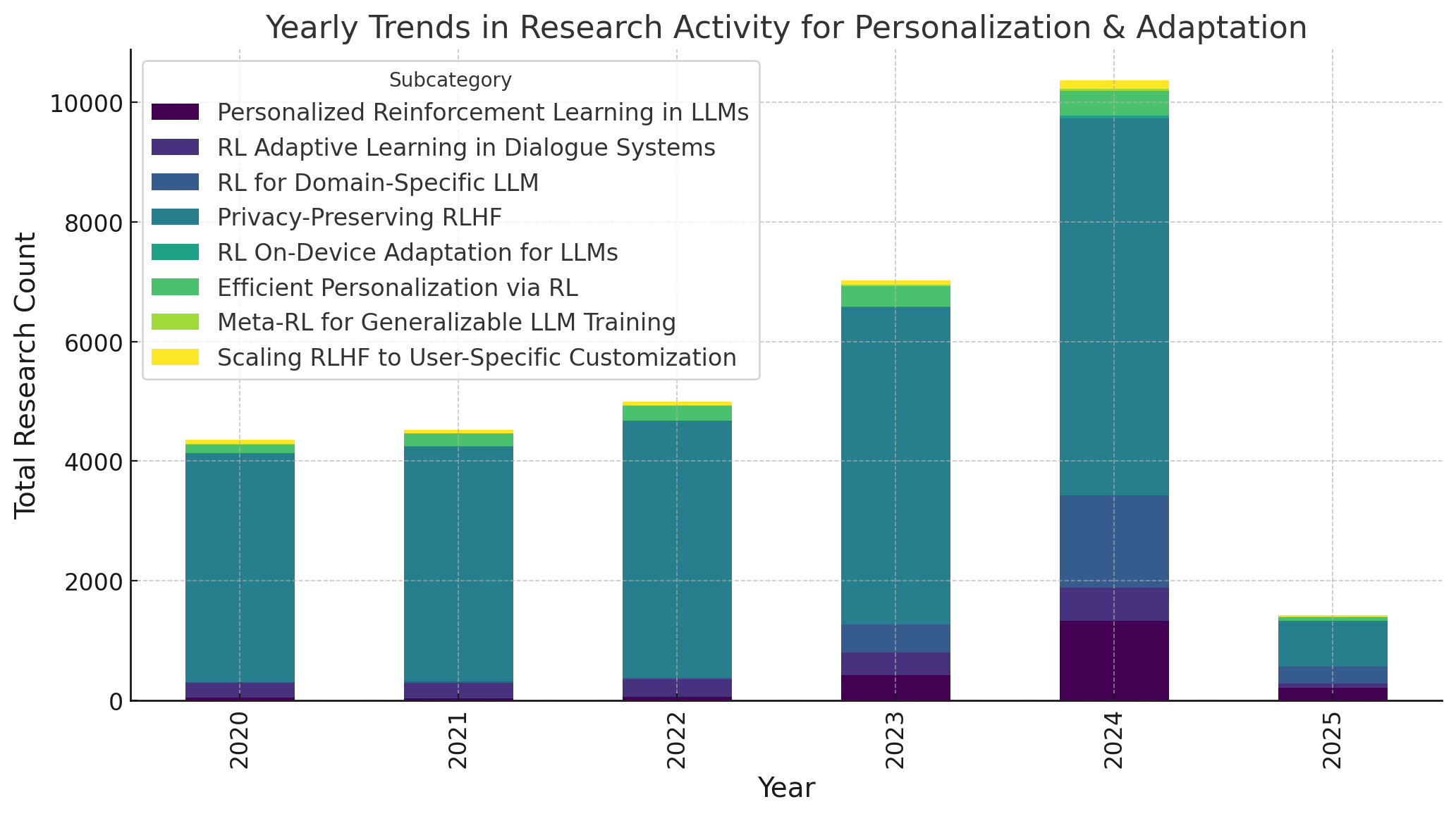}
        \caption{\small Personalization and Adaptation focus on Privacy-Preserving \rlhf. On-device adaptation remains a challenge.}
        \label{fig:personalization}
    \end{subfigure}
    \hfill
    \begin{subfigure}[b]{0.32\textwidth}
        \centering
        \includegraphics[width=\textwidth]{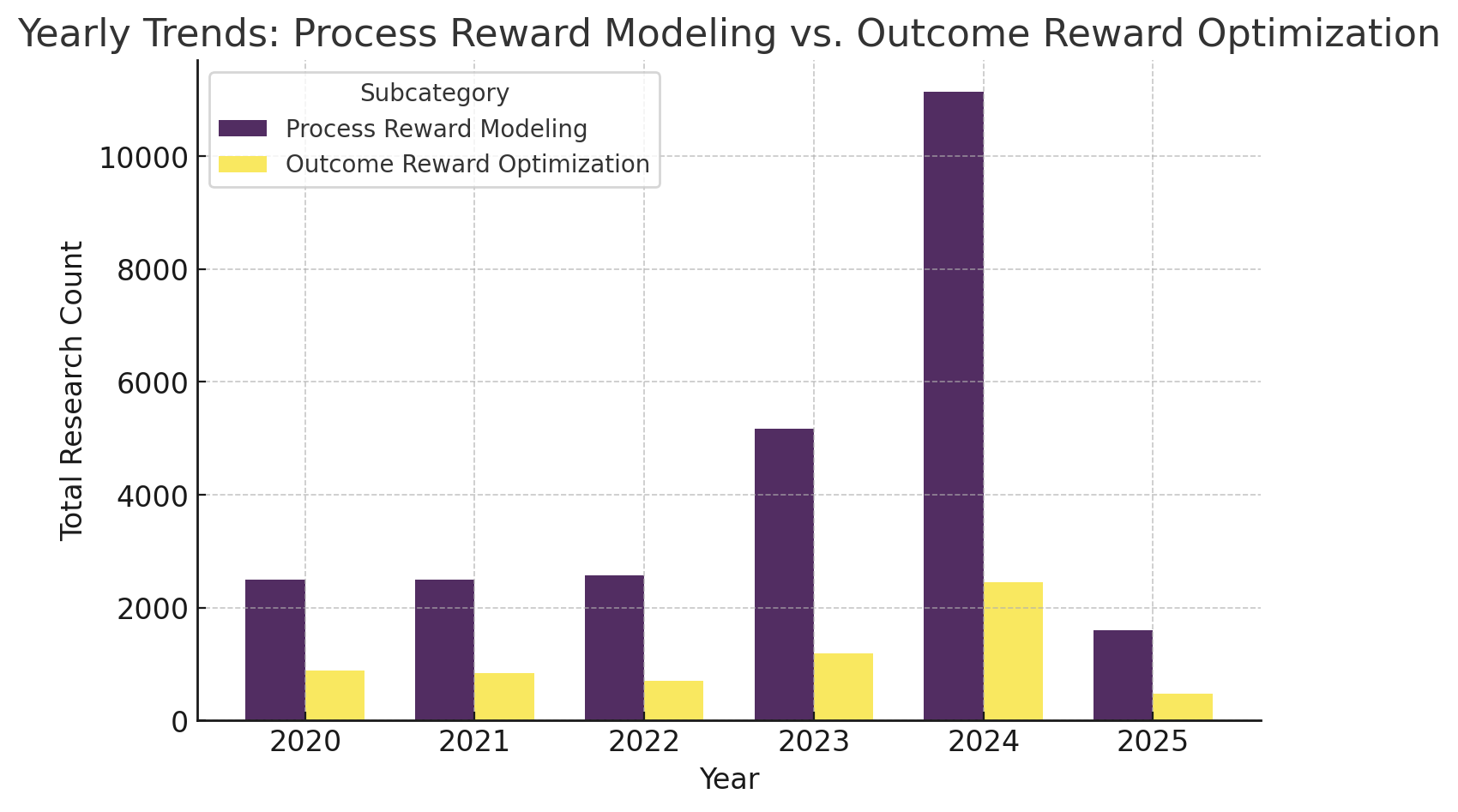}
        \caption{\small Process Reward Modeling dominates Outcome-Based Optimization, favoring iterative strategies for RL-based LLMs.}
        \label{fig:process_vs_outcome}
    \end{subfigure}
    \caption{Yearly Trends in \rl specific post-training methods for LLMs and emerging research directions.}
    \label{fig:combined_trends}
\end{figure*}

We gathered all papers related to post-training methods in \llms and analyzed their trends, as shown in Figure \ref{fig:combined_trends}. Application of \rl techniques \cite{rlhf,rafailov2024direct,guo2025deepseek} for refining the LLMs have a noticeable increase in prominence since 2020 (Figure~\ref{fig:advanced_rl}), emphasizing the demand for \textbf{interactive approaches} such as human-in-the-loop \cite{pan2024human, Selfcritiquing} reinforcement and scalability \cite{geiping2025scalingtesttimecomputelatent,gao2023scaling,kaplan2020scaling}. At the same time, \textbf{reward modeling} \cite{roy2016solvinggeneralarithmeticword, rewardingprogress, lambert2024rewardbench} (Figure~\ref{fig:reward_modeling}) has seen a steady rise in interest due to the emergence of self-rewarding language models, yet the field still struggles with \textbf{reward hacking} \cite{jinnai2024regularized,Chen2024ODINDR} and the design of robust \cite{Liu2024RRMRR}, failure-aware reward functions beyond reward hacking \cite{Wang2025BeyondRH}. 
\textbf{Decoding and search} (Figure~\ref{fig:decoding}) methods include tree-of-thoughts \cite{yao2024tree} and Monte Carlo \cite{browne2012survey, sprueill2023monte} strategies aiming to enhance model reasoning through iterative self-critique \cite{selfrefine,Selfcritiquing,ye2024self}, but these techniques also demand reliable uncertainty estimators to prevent excessive computational overhead \cite{fu2022complexity,geiping2025scalingtesttimecomputelatent}. 
Safety \cite{ji2024beavertails,johnson2022metacognition,openai2024o3}, robustness \cite{Yan2024RewardRobustRI}, and interpretability \cite{yu2022explainable, amini2019mathqainterpretablemathword,Chen2023DrivingWL} have likewise become central concerns (Figure~\ref{fig:safety}), motivating the development of bias-aware \cite{poulain2024bias, fan2024biasalert} and uncertainty-aware \cite{li2023coannotating} RL methods beyond correlation with human uncertanity \cite{elangovan2024beyond} that safeguard user trust and prevent adversarial attacks. Another crucial area involves \textbf{personalization} \cite{dong2024can,wang2023learning} and \textbf{adaptation} \cite{zhang2023adaptive} (Figure~\ref{fig:personalization}), where efforts to tailor LLMs for specific domains must be balanced against risks to privacy \cite{du2024privacyfinetuninglargelanguage}, particularly when enterprise data or sensitive personal information is involved. 

In parallel, \textbf{process} \cite{yuan2024implicitprm,Ma2023EurekaHR} vs.\ \textbf{outcome reward optimization} \cite{Havrilla2024GLoReWW} (Figure~\ref{fig:process_vs_outcome}) remains an open question: while process-based rewards help guide incremental improvements, outcome-focused metrics are simpler but may not capture crucial intermediate decision-making steps. Beyond reward structure, \textbf{fine-tuning} \llms on new tasks still encounter issues like \textbf{catastrophic forgetting} \cite{URLoRA} and potential data leakage \cite{Fu2023MMEAC,Wang2023PandaLMAA}, underscoring the need for parameter-efficient methods \cite{hu2021lora} and privacy-preserving strategies such as differential privacy \cite{Sun2024ImprovingLI} and federated learning \cite{he2024fedeval}. Human feedback, while central to alignment, is inherently costly and limited in scope; methods like \textbf{Constitutional AI} \cite{bai2022constitutional} and \texttt{RLAIF} \cite{lee2023rlaif} seek to automate parts of this oversight, though they introduce fresh concerns about bias calibration \cite{park2024offsetbias} and model self-consistency \cite{wang2023selfconsistency}. 
Finally, test-time scaling \cite{geiping2025scalingtesttimecomputelatent} and \textbf{dynamic reasoning} \cite{lu2023dynamicpromptlearningpolicy} frameworks pose further challenges: models must learn when to allocate more computation for complex queries, how to adapt verification modules \cite{generativeverifiers} efficiently, and how to maintain robust performance even when facing adversarial inputs. These converging research directions—spanning reward modeling, decoding strategies, interpretability, personalization, and safe fine-tuning—highlight the multifaceted role of \rl in \llms and collectively shape the future trajectory of large-scale language model development. Below, we delve into some of these directions in greater detail.

\noindent\textbf{Fine-tuning challenges.}
Fine-tuning remains one of the most direct post-training methods to adapt \llms to specific tasks or domains, yet it faces several open challenges. One fundamental issue is catastrophic forgetting – when updating an \llm on new data causes it to lose or degrade previously learned capabilities. Even advanced \texttt{PEFT} methods like \lora  \cite{hu2021lora}, which greatly reduce the number of trainable weights, do not fully solve this problem \cite{URLoRA}. 
Future work can explore better continual learning strategies and regularization techniques so that models can acquire new skills without erasing old ones.  
For example, new fine-tuning algorithms (e.g. \texttt{CURLoRA} \cite{URLoRA}) explicitly aim to stabilize training and preserve prior knowledge while adding new tasks.
Promising research directions include curriculum-based fine-tuning \cite{promptcurri} (introducing new facts gradually or in context with known facts) and hybrid training that combines retrieval or external knowledge bases. For instance, rather than solely adjusting the model’s weights, one could fine-tune \llms to consult a knowledge repository or perform tool use (such as database queries or computations) when faced with queries outside their original training distribution \cite{yu2025rankrag,lin2023ra}. This retrieval-augmented fine-tuning \cite{zhang2024raft} could let models incorporate fresh information at inference time, reducing the need to overwrite their internal weights with new facts. Another approach is training models to explicitly represent uncertainty about new knowledge, thereby enabling them to say `I don’t know' or defer to an external source if a query concerns content not seen in pre-training. By blending weight updates with external knowledge integration, future fine-tuned \llms will maintain higher factual accuracy and lower hallucination rates on emerging information.

\noindent\textbf{Safe Fine-tuning.}
From an ethical and safety perspective, fine-tuning raises important open research questions. Fine-tuning data often contains sensitive or proprietary information \cite{du2024privacyfinetuninglargelanguage}, which can lead to privacy risks if the model memorizes and later regurgitates that data. A recent comprehensive survey \cite{huang2024harmful} highlights vulnerabilities in the fine-tuning stage, such as membership inference attacks (detecting if a specific record was in the fine-tuning set) and data extraction (recovering parts of the fine-tuning data from the model’s outputs). Mitigating these risks is an open problem: methods like differential privacy fine-tuning \cite{Sun2024ImprovingLI} (adding noise to the weight updates) and federated fine-tuning (where data never leaves user devices and only aggregated updates are sent to the model) are being actively explored. However, these methods often come at the cost of model utility or require careful calibration to avoid degrading performance.

\noindent\textbf{Limitations of Human Feedback.}
Human feedback is costly and subjective. One promising avenue to address the limitations of human feedback is using AI feedback and automation to assist or replace human evaluators. Constitutional AI \cite{bai2022constitutional}, introduced by Anthropic, is a notable example: instead of relying on extensive human feedback for every harmful or helpful behavior, the model is guided by a set of written principles (a `constitution') and is trained to critique and refine its own responses using another AI model as the judge \cite{bowman2022measuring}.
Emerging directions here include  \texttt{RLAIF} \cite{lee2023rlaif} and other semi-automated feedback techniques \cite{hollmann2023llms}: using strong models to evaluate or guide weaker models, or even having multiple AI agents debate a question and using their agreement as a reward signal \cite{motwani2024malt,estornell2024acc}. 
Such AI-aided feedback could vastly scale the tuning process and help overcome the bottleneck of limited human expert time. However, it raises new theoretical questions: how do we ensure the AI judge is itself aligned and correct? There is a risk of feedback loops or an echo chamber of biases if the automated preferences are flawed. 
An open gap is the creation of robust AI feedback systems that are calibrated to human values (perhaps periodically `grounded' by human oversight or by a diverse set of constitutional principles). 
The blending of human and AI feedback in a hierarchical scheme could provide a scalable yet reliable \rl paradigm for \llms.

\noindent\textbf{Test-time scaling challenges.}
Open challenges in \texttt{TTS} revolve around how to orchestrate the inference-time processes efficiently and reliably. 
A key question is how much computing is enough for a given query, and how to determine this on the fly? Using less resources can result in mistakes, but using too much is inefficient and could introduce inconsistencies. 
Recent research by Snell et al. \cite{snell2024scaling} tackled it by proposing a unified framework with a `Proposer' and a `Verifier' to systematically explore and evaluate answers. 
In their framework, the Proposer (usually the base \llm) generates multiple candidate solutions, and the Verifier (another model or a heuristic) judges and selects the best. 
The optimal strategy can vary by problem difficulty: for easier queries, generating many answers in parallel and picking the top might be sufficient, whereas for harder problems, sequential, step-by-step reasoning with verification at each step works better. 
An important future direction is building adaptive systems where the \llm dynamically allocates computation based on an estimate of the question’s complexity. 
This idea connects to meta-cognition in AI \cite{johnson2022metacognition}, enabling models to have a sense of what they don’t know or what deserves more thought. 
Developing reliable confidence metrics or difficulty predictors for \llms is an open research area, but progress here would make \texttt{TTS} far more practical i.e., the model would only `slow down and think' when necessary, much like a human spending extra time on a hard problem.
Additionally, By reframing inference-time scaling as a probabilistic inference problem and employing particle-based Monte Carlo methods \cite{puri2025probabilistic}, the small models achieved o1 level accuracy in only 32 rollouts, a 4–16x improvement in scaling efficiency across various mathematical reasoning tasks.
Recent study \cite{luo2024improve} shows distilling test-time computations into synthetic training data creates synergistic pretraining benefits which can also be further explored.

\noindent\textbf{Reward Modeling and Credit Assignment.}
Current \rl approaches suffer from reward misgeneralization, where models over-optimize superficial proxy metrics rather than genuine reasoning quality. The sparse nature of terminal rewards in multi-step tasks increases credit assignment challenges, particularly in long-horizon reasoning scenarios. Traditional methods like \dpo require inefficient pairwise preference data and fail to utilize failure trajectories effectively.
Hybrid reward models can be investigated by integrating process supervision with outcome-based rewards using contrastive stepwise evaluation \cite{shen2024improving}. This approach enables a more granular assessment of intermediate decision-making steps while aligning with long-term objectives. Recent work \cite{oreo} suggests step-level policy optimization could improve value function accuracy while maintaining safety constraints.
Dynamic credit assignment mechanisms can be explored through temporal difference learning adapted for transformers \cite{ma2021long,pignatelli2023survey}. Such adaptations may enhance the model’s ability to capture long-range dependencies and optimize reward propagation over extended sequences.
Failure-aware training strategies can be developed by incorporating negative examples into the \rl loop via adversarial data augmentation \cite{zhang2021generalizationreinforcementlearningpolicyaware}. This can improve model robustness by systematically exposing it to challenging scenarios and encouraging more resilient policy learning.

\noindent\textbf{Efficient \rl Training and Distillation.}
Current \rl methods for \llms require prohibitive computational resources~\cite{ahmadian2024back} while often underperforming knowledge distillation techniques \cite{wu2025survey}. 
This inefficiency limits scalability and practical deployment, as distilled models frequently surpass \rl-trained counterparts despite requiring less training overhead. Additionally, pure \rl approaches struggle to balance language quality with reasoning improvement \cite{zhao2023survey,wu2025survey}, creating a performance ceiling.

The development of hybrid frameworks that initialize \rl policies with distilled knowledge from large models, combining the exploratory benefits of \rl with the stability of supervised learning is an interesting direction. Similarly, curriculum sampling strategies that progressively increase task complexity while using distillation to preserve linguistic coherence can also help. \texttt{PEFT} methods \cite{hu2021lora} can be leveraged during \rl updates to maintain base capabilities while enhancing reasoning.

\insightbox{Integration: Combining PRM-guided tree search with online distillation achieves 4× efficiency gains over baseline methods, while maintaining 94\% solution accuracy on MATH dataset.}

\noindent\textbf{Privacy-Preserving Personalization.}
Customizing models for enterprise and individual use cases raises the risk of exposing private training data through memorization, making privacy-preserving \cite{du2024privacyfinetuninglargelanguage} adaptation essential. Promising solutions include homomorphic instruction tuning \cite{lee2024hetalefficientprivacypreservingtransfer}, which processes encrypted user queries while maintaining end-to-end encryption during inference; differential privacy via reward noising \cite{Wei2024DistributedDP}, which introduces mathematically bounded noise into \rlhf preference rankings during alignment; and federated distillation, which aggregates knowledge from decentralized user-specific models without sharing raw data. 

\noindent\textbf{Collaborative Multi-Model Systems.}
As single-model \cite{zhang2024agent,ma2024agentboard,zhang2024aflow} scaling approaches physical limits, alternative paradigms such as multi-agent \llm collaboration \cite{le2024multi,owens2024multi,sanh2021multitask} become necessary. 
Researchers are investigating emergent communication protocols that train models to develop lossy compression ``languages'' for inter-model knowledge transfer such as GenAINet \cite{Zou2024GenAINetEW}, robust ensembles where stress-test induced specialization drives automatic division of problem spaces based on failure analysis \cite{lee2020adaptivestresstestingfinding}, and gradient-free synergy learning through evolutionary strategies designed to discover complementary model combinations without relying on backpropagation \cite{baydin2022gradientsbackpropagation}.

\noindent\textbf{Multimodal RL Integration.}
Multimodal reinforcement learning \cite{Liu2024ArondightRT, li2025imaginereasoningspacemultimodal,kim2024aligning} faces the obstacle of a combinatorial state explosion, especially in contexts exceeding 128k tokens. Pioneering methods to overcome this include hierarchical attention frameworks that employ modality-specific policies with cross-attention gating \cite{Ebrahimi2024CROMECA}, adaptive truncation strategies that compress context while preserving critical reasoning segments \cite{Xia2025TokenSkipCC}, and flash curriculum approaches that leverage self-supervised \cite{Ma2023LeveragingSP,Xi2024TrainingLL,Lee2024AffordanceGuidedRL} complexity prediction to facilitate progressive multimodal integration. 

\noindent\textbf{Efficient \rl Training.}
Efficient \rl training paradigms continue to be a critical research frontier as current methods exhibit significant sample inefficiency and computational overhead. Addressing issues like the overthinking \cite{Xu2024RejectionIR,Chen2024DoNT} phenomenon, where excessive reasoning chains waste valuable computation \cite{team2025kimi}, requires approaches such as partial rollout strategies \cite{kemmerling2024beyond}, adaptive length penalty mechanisms employing learned compression transformers, and hybrid architectures that combine MCTS with advanced \rl optimizers. These innovations are essential for scaling \rl to long-context tasks while minimizing wasted computational resources.
\warningbox{Overthinking Phenomenon: Analysis reveals 22\% wasted computation is in reasoning chains exceeding optimal reasoning length.}
\rl methods exhibit sample inefficiency and computational overhead, particularly when scaling to contexts exceeding 128k tokens. The `overthinking' phenomenon, where models generate excessively long reasoning chains, further reduces token efficiency and increases deployment costs \cite{Li2024PersonalLA}.
Investigate partial rollout strategies with flash attention mechanisms for long-context processing. Develop length penalty mechanisms using learned compression transformers for iterative long2short distillation. Hybrid architectures combining \texttt{MCTS}~\cite{xie2024monte} with \grpo \cite{shao2024deepseekmath} could enable better exploration-exploitation tradeoffs. Parallel work by Xie et. al.~\cite{xie2024monte} demonstrates promising results through adaptive tree search pruning.
Several open challenges persist in the field. Uncertainty propagation remains problematic as current confidence estimators add approximately 18\% latency overhead, while catastrophic forgetting rresults in a degradation of 29\% of base capabilities during \rl fine-tuning \cite{li2024revisitingcatastrophicforgettinglarge}. Moreover, benchmark saturation is an issue, with \texttt{MMLU} scores correlating poorly (r = 0.34) with real-world performance \cite{alzahrani2024benchmarkstargetsrevealingsensitivity}. \warningbox{Adversarial Vulnerabilities: Stress tests reveal a high success rate on gradient-based prompt injections.}

%% file: content/conclusion.tex
\section{Conclusion}

This survey and tutorial provides a systematic review of post-training methodologies for \llms, focusing on fine-tuning, reinforcement learning, and scaling. We analyze key techniques, along with strategies for improving efficiency and alignment with human preferences. Additionally, we explore the role of  \rl in enhancing \llms through reasoning, planning, and multi-task generalization, categorizing their functionalities within the agent-environment paradigm.  
Recent advancements in reinforcement learning and test-time scaling have significantly improved \llms reasoning capabilities, enabling them to tackle increasingly complex tasks. By consolidating the latest research and identifying open challenges, we aim to guide future efforts in optimizing \llms for real-world applications.